\documentclass{article}
\PassOptionsToPackage{numbers,sort&compress}{natbib}

\usepackage[preprint]{neurips_2026}

\usepackage{microtype} 
\usepackage[utf8]{inputenc}				
\usepackage[T1]{fontenc}				
\usepackage{graphicx} 
\usepackage{hyperref}
\usepackage{amsmath, amssymb, amsthm, mathtools, bm}
\usepackage[dvipsnames]{xcolor}
\usepackage{caption}
\usepackage{url}
\usepackage{booktabs} 
\usepackage[labelformat=simple]{subcaption}
\usepackage{nicefrac}					
\usepackage{multirow}
\usepackage{enumitem}

\graphicspath{{./figures/}}
\DeclareGraphicsExtensions{.pdf}

\hypersetup{
	hidelinks,
    colorlinks=true,
    linkcolor=Blue,
    filecolor=Blue,
    citecolor=Blue,
    urlcolor=Blue,
    pdftitle={Learning When to Adapt},
}

\theoremstyle{plain}

\theoremstyle{definition}

\theoremstyle{remark}

\newcommand{\norm}[1]{\left\lVert#1\right\rVert_2}

\newcommand\rb[1]{\left(#1\right)}
\renewcommand\sb[1]{\left[#1\right]}

\renewcommand\aa[0]{\mathbf{a}}
\newcommand\bb[0]{\mathbf{b}}

\newcommand\xx[0]{\mathbf{x}}
\newcommand\yy[0]{\mathbf{y}}

\renewcommand\AA[0]{\mathbf{A}}
\newcommand\BB[0]{\mathbf{B}}

\newcommand\II[0]{\mathbf{I}}
\newcommand\UU[0]{\mathbf{U}}
\newcommand\VV[0]{\mathbf{V}}
\newcommand\WW[0]{\mathbf{W}}

\newcommand\MM[0]{\mathbf{M}}

\newcommand\GG[0]{\mathbf{G}}

\newcommand{\ours}{\textsc{DIS}e\textsc{L}}
\newcommand{\roberta}{\textsc{RoBERT}a}
\newcommand{\lora}{\textsc{LoRA}}

\newcommand{\llama}{\textsc{Llama}}
\newcommand{\mistral}{\textsc{Mistral}}

\newcommand{\ww}{\mathbf{w}}
\newcommand{\zeroVec}{\boldsymbol{0}}
\newcommand{\R}{\mathbb{R}}
\newcommand{\E}{ \mathbb{E} }

\newcommand{\bmu}{\boldsymbol{\mu}}
\newcommand{\btheta}{\boldsymbol{\theta}}
\newcommand{\CovMat}{\boldsymbol{\Sigma}}
\newcommand{\rank}{\operatorname{rank}}
\newcommand{\trace}{\operatorname{Tr}}
\newcommand{\bDelta}[0]{\boldsymbol{\Delta}}

\title{Learning When to Adapt}

\author{%
  Ali Zindari$^{1,2}$ ~ Xiaowen Jiang$^{1,2}$ ~ Rotem Mulayoff$^{2}$ ~ Sebastian U.\ Stich$^{2}$\\%
  $^{1}$Universit\"at des Saarlandes ~ $^{2}$CISPA Helmholtz Center for Information Security\\%
  Saarbr\"ucken, Germany\\%
  \texttt{\{ali.zindari,xiaowen.jiang,rotem.mulayoff,stich\}@cispa.de}%
}

\begin{document}

\maketitle

\begin{abstract}
    Low-rank adaptation (\lora{}) is a widely used parameter-efficient fine-tuning method, yet its learned correction is static: the same low-rank update is applied to every input. This input-agnostic approach creates an inevitable compromise between adapting to the fine-tuning distribution and preserving pre-trained behavior on inputs outside that distribution, contributing to \emph{catastrophic forgetting}. We introduce \ours{} (\textbf{D}ynamic \textbf{I}nput-\textbf{Se}nsitive \textbf{L}oRA), which augments \lora{} modules with lightweight input-dependent gates over individual rank-one components. The gating mechanism is designed to preserve the pre-trained model’s behavior by default, while training learns to activate selected components that reduce the fine-tuning loss. \ours{} adds only a small number of parameters and preserves the low-rank structure. Across \roberta{} on GLUE, and \llama{} and \mistral{} models fine-tuned for mathematical reasoning and code generation, \ours{} reduces forgetting relative to LoRA and related variants while maintaining competitive fine-tuning accuracy. In addition, the learned gate activations provide an interpretable diagnostic view of which layers and rank components are most activated during fine-tuning, giving insight into where task-specific adaptation is concentrated. Code available at: \url{https://github.com/alizindari/DISeL}.
\end{abstract}

\section{Introduction}\label{sec:introduction}
Fine-tuning (FT) has become a central stage in the deployment of large language models (LLMs) where pre-trained models with general knowledge are specialized for instruction-following~\citep{ouyang2022training, wei2022finetuned}, mathematical reasoning~\citep{yu2024metamath, yue2024mammoth}, code generation~\citep{roziere2023code, luo2024wizardcoder}, and other downstream tasks. The most direct approach is \emph{full fine-tuning}, which updates all parameters of the pre-trained model without constraint. Parameter-efficient fine-tuning (PEFT) methods instead restrict updates to a smaller, structured subspace. The most popular approach is Low-Rank Adaptation (\lora{})~\citep{hu2022lora}, which freezes the pre-trained weights and learns a low-rank correction using matrix factorization.

Despite achieving strong performance on their target tasks, fine-tuned models often exhibit degradation in their broader capabilities acquired during pre-training~\citep{luo2025empirical, dong2024abilities}. This phenomenon, known as \emph{catastrophic forgetting}, has been frequently studied in continual learning~\citep{wang2024comprehensive, liang2025gated, wang2022s, wang2023hierarchical, mcdonnell2023ranpac} and remains a key challenge in modern LLM fine-tuning. The extent of forgetting depends on the choice of fine-tuning method. Typically, full fine-tuning incurs substantial performance degradation outside the fine-tuning domain~\citep{biderman2024lora, luo2025empirical}, whereas \lora{} mitigates this effect but does not eliminate it~\citep{biderman2024lora, shuttleworth2025lora}.

The main reason for forgetting in these FT methods is that the learned modification is input-agnostic. Full fine-tuning, \lora{}, and its variants such as \textsc{DoRA}~\citep{liu2024dora} and \textsc{AdaLoRA}~\citep{zhang2023adalora} only differ in parameterization. Yet, they all apply a constant modification across the entire input space, regardless of whether correction is required. As a result, the same adaptation affects both fine-tuning-domain inputs, where correction is needed, and pre-training-domain inputs, where the original mapping should be preserved. This leads to an inherent compromise between adaptation and retention.

Existing input-dependent adaptations are primarily designed to mitigate interference across tasks or among multiple \lora{} modules. For instance, mixture-of-\lora{} approaches~\citep{wu2024mixture, buehler2024x, tian2024hydralora} learn routing or gating mechanisms that select or combine multiple \lora{} modules based on the input. The closest to this work is \textsc{Gated LoRA}~\citep{eom2025gatedlora}, which introduces for each rank-one-component, an input-dependent gate within a single adapter. However, its primary goal is to reduce inter-task interference in multi-task fine-tuning, rather than to preserve pre-trained behavior. None of the current methods explicitly target catastrophic forgetting, nor are they designed to make the adapter inactive on inputs outside the fine-tuning distribution, a key property for preserving pre-trained capabilities.

To this end, we propose \ours{} (\textbf{D}ynamic \textbf{I}nput-\textbf{Se}nsitive \textbf{L}o\textbf{RA}), which preserves the low-rank parameterization of \lora{} but makes the correction input-dependent and active only when necessary. The key idea is to multiply each rank-one component by an input-dependent learnable gate whose output lies in $[0,1]$. The gates are effectively pre-set to zero in initialization, meaning that \ours{} begins training with the pre-trained model. During training, gates are activated only when doing so reduces the fine-tuning loss, leading to updates that are applied selectively across inputs rather than uniformly. The computational overhead for training relative to existing methods is negligible, while the inference cost matches that of \lora{}, making \ours{} a practical alternative.

\paragraph{Contributions.} In this work, we introduce an FT method that minimizes forgetting while maintaining strong performance on the fine-tuning task. Specifically, we make the following contributions.
\begin{itemize}[leftmargin=5mm]
    \item We identify the input-agnostic nature of full fine-tuning/\lora{} modifications as a contributing factor in catastrophic forgetting, and illustrate this in a minimal linear regression setting (Sec.~\ref{sec:background}).
    \item We introduce \ours{}, which augments each \lora{} module with lightweight, input-dependent gates that activate only when needed, at negligible parameter and compute overhead (Sec.~\ref{sec:method}).
    \item We conduct experiments on \llama{} and \mistral{} models fine-tuned for mathematical reasoning and code generation. \ours{} consistently reduces catastrophic forgetting compared with \lora{} and its variants, while maintaining competitive in-domain accuracy (Sec.~\ref{sec:experiments}).
    \item The learned gate activations reveal which layers are most engaged during fine-tuning, providing direct insight into where task-specific knowledge is localized within the network (Sec.~\ref{sec:interpretability}).
\end{itemize}

\section{Background and motivation}\label{sec:background}

\textbf{Fine-tuning.} Let $\WW_0$ denote the parameters of a pre-trained model. Standard fine-tuning methods learn an additive update of the form
\begin{equation}
    \label{eq:FT-InputIndependent}
    \WW = \WW_0 + \bDelta .
\end{equation}
Depending on the method, $\bDelta$ can have a specific structure.
Full FT~\citep{ouyang2022training,devlin2019bert}
updates the entire parameter set $\WW$ without constraints, incurring high memory overhead and often increasing susceptibility to forgetting~\citep{biderman2024lora}. 
In contrast, parameter-efficient fine-tuning (PEFT) restricts updates to a small subset of parameters while freezing the rest, including adapters and prompt-based techniques. For instance, \lora{}~\citep{hu2022lora} and its variants restrict $\bDelta$ to a low-rank correction using matrix factorization.

After fine-tuning, these methods apply the learned modification to every input, \emph{i.e.}, they are input-agnostic. Although the resulting functional change can still vary with the input through the model's activations, the update itself has no mechanism for selectively turning off on inputs where the pre-trained behavior should be preserved. As a result, both fine-tuning-domain inputs, where modification is beneficial, and pre-training-domain inputs, where the original behavior should be preserved, are affected equally. To study this source of forgetting, we analyze a minimal linear regression model in which both pre-training and fine-tuning data remain available throughout training. We show that even in this favorable scenario, any fixed additive update of the form~\eqref{eq:FT-InputIndependent} must compromise between fitting the fine-tuning data and preserving the pre-trained mapping.

\begin{figure}[t]%
    \begin{subfigure}[t]{0.5\linewidth}%
        \includegraphics[width=\linewidth]{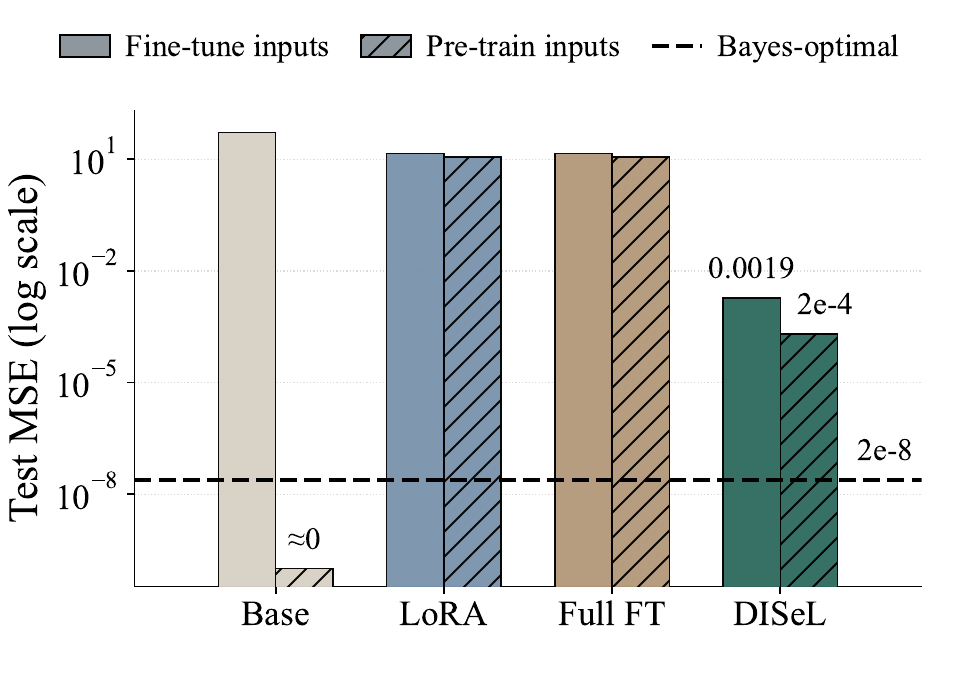}
        \caption{Test MSE}
        \label{fig:toy-mse}
    \end{subfigure}%
    \begin{subfigure}[t]{0.5\linewidth}%
        \includegraphics[width=\linewidth]{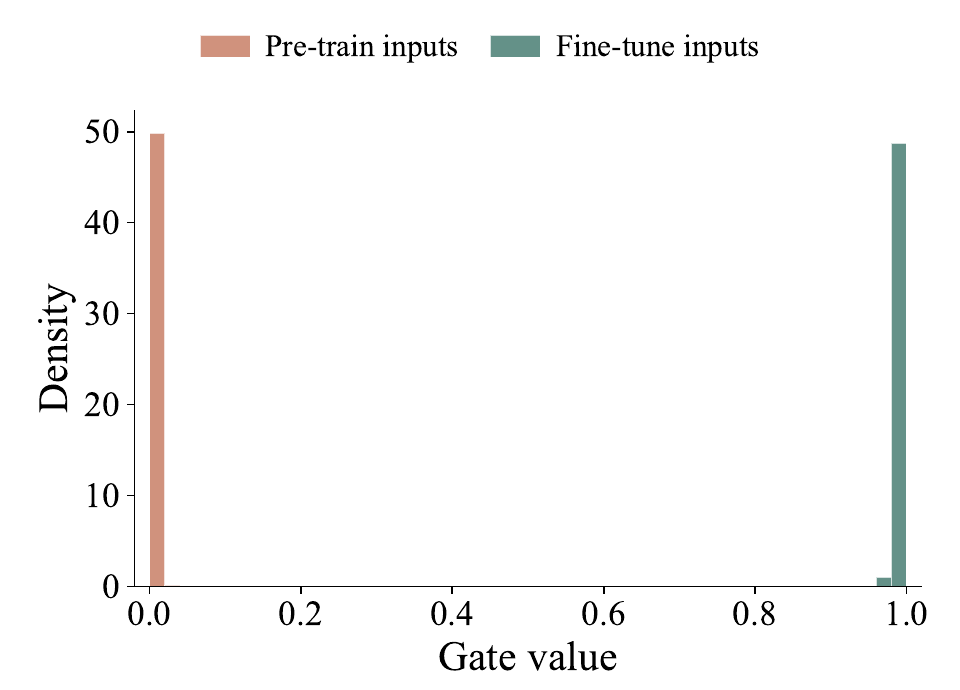}
        \caption{Gate histogram}
        \label{fig:toy-gates}
    \end{subfigure}
    \caption{{\bf Selective adaptation toy example.} 
    We minimize the loss in~\eqref{eq:fixed-loss} on inputs drawn from a symmetric mixture of two Gaussian populations (see Appendix~\ref{sec:appendix-toy} for details). Panel~\protect\subref{fig:toy-mse} shows the test MSE on fine-tuning and pre-training inputs. Here, \lora{} and full FT reach the fixed-correction tradeoff, whereas \ours{} achieves much lower error on both domains, approaching the Bayes-optimal error floor (dashed line). Panel~\protect\subref{fig:toy-gates} plots the activation of the learned gates in \ours{}. We see that the activations concentrate near $1$ for fine-tuning inputs and near $0$ for pre-training inputs, confirming that the update is activated only when adaptation is needed.}
    \label{fig:toy}
\end{figure}

\textbf{A minimal model of adaptation and retention.} Let $p_{\text{ft}}$ and $p_{\text{pt}}$ denote the fine-tuning and pre-training data distributions, and let $\WW_0 \in \R^{d_y \times d_x}$ denote the pre-trained model. Each training pair~$(\xx, \yy)$ is sampled from the symmetric mixture of the two populations as follows:
\begin{equation}
    \xx \sim \begin{cases}
        p_{\text{ft}} & \text{w.p. }~  \nicefrac{1}{2} , \\
        p_{\text{pt}} & \text{w.p. }~ \nicefrac{1}{2},
    \end{cases}
    \qquad
    \yy = \begin{cases}
        (\WW_0 + \MM)\xx & \text{when } \xx \sim p_{\text{ft}}, \\
        \WW_0\xx & \text{when } \xx \sim p_{\text{pt}},
    \end{cases}
\end{equation}
where $\MM \in \R^{d_y \times d_x}$ is an arbitrary task-specific matrix.
This setting is strictly easier than sequential fine-tuning since both populations are available throughout training. Adaptation is therefore required on $p_{\text{ft}}$,
while the original mapping should be preserved on $p_{\text{pt}}$, leading to the following objective:
\begin{equation}\label{eq:fixed-loss}
    \mathcal{L}(\bDelta) =
    \tfrac12 \E_{p_{\text{ft}}}\sb{\norm{(\bDelta - \MM)\xx}^2}
    +
    \tfrac12 \E_{p_{\text{pt}}}\sb{\norm{\bDelta\xx}^2}.
\end{equation}
Let 
$\CovMat^{\text{ft}}_{\xx\xx}$ and $\CovMat^{\text{pt}}_{\xx\xx}$ be the second-moment matrices under the distributions $p_{\text{ft}}$ and $p_{\text{pt}}$. To expose the conflict in its simplest form, suppose $\CovMat^{\text{ft}}_{\xx\xx} = \CovMat^{\text{pt}}_{\xx\xx} =: \CovMat_{\xx\xx}$. Thus, the unconstrained optimum is
\begin{equation}
    \bDelta^\star = \tfrac12\MM,
    \qquad
    \mathcal{L}(\bDelta^\star) = \tfrac14\trace\rb{\MM \CovMat_{\xx\xx} \MM^\top}.
    \label{eq:half-m}
\end{equation}
In this linear setting, full FT corresponds to optimizing an unconstrained fixed matrix $\bDelta$, while \lora{} further restricts $\bDelta$ to a low-rank factorization.
Imposing additional constraints on $\bDelta$, such as a low-rank factorization, can only increase the loss, so $\mathcal{L}(\bDelta^\star)$ is a lower bound for any fixed-update method. However, even this best fixed update achieves neither full adaptation on $p_{\text{ft}}$ nor zero forgetting on $p_{\text{pt}}$: it applies only half of the desired correction on fine-tuning inputs and a nonzero half-correction on pre-training inputs. This reflects a structural conflict arising from using the same fixed matrix across both populations. Figure~\ref{fig:toy} confirms this empirically: both \lora{} and full FT incur non-negligible MSE on each population, illustrating the tradeoff. 

This limitation can be mitigated by allowing the correction to depend on the input. In this regression model, the optimal input-dependent correction takes the form (see derivation in Appendix~\ref{sec:appendix-toy}):
\begin{equation}
    f^\star(\xx) = \pi_{\mathrm{ft}}(\xx)\MM\xx,
    \qquad
    \pi_{\mathrm{ft}}(\xx)
    =
    \frac{p_{\mathrm{ft}}(\xx)}
    {p_{\mathrm{ft}}(\xx)+p_{\mathrm{pt}}(\xx)} \in [0,1].
\end{equation}
Here $\pi_{\mathrm{ft}}(\xx)$ is the posterior probability that an input with value $\xx$ was drawn from the fine-tuning population under the symmetric mixture. Thus, the update is large on inputs that are likely under the fine-tuning distribution and small on inputs that are likely under the pre-training distribution. This correction achieves the minimum possible squared-error loss for the problem; in the special case where the two distributions have disjoint support, $\pi_{\mathrm{ft}}(\xx)$ is an indicator and the loss becomes zero.

When the two populations are multivariate Gaussian with the same covariance $\CovMat$ and different means $\bmu_{\mathrm{ft}}\neq \bmu_{\mathrm{pt}}$, the log density ratio is affine in $\xx$, and the optimal correction thus admits the closed form
\begin{equation}\label{eq:bayes-gaussian}
    f^\star(\xx) =\sigma\rb{\ww_g^\top \xx + b_g}\MM\xx, 
\end{equation}
where $\sigma(x) \coloneqq {1}/{(1+e^{-x})} \in [0,1]$ for 
$x \in \R$ is the sigmoid function, and 
\begin{equation}
    \ww_g \coloneqq \CovMat^{-1}(\bmu_{\text{ft}} - \bmu_{\text{pt}}) \in \R^d,
    \qquad
    b_g \coloneqq \tfrac12 \left(
    \bmu_{\text{pt}}^\top\CovMat^{-1}\bmu_{\text{pt}} -\bmu_{\text{ft}}^\top\CovMat^{-1}\bmu_{\text{ft}}
    \right) \in \R.
\end{equation}
The full derivations can be found in Appendix~\ref{sec:appendix-toy}.

From~\eqref{eq:bayes-gaussian}, we see that the optimal correction in this case is a \emph{sigmoid of an affine transformation of $\xx$}, applied to the fine-tuning update $\MM\xx$. This motivates the introduction of an input-dependent sigmoid gate $\sigma(\ww_g^\top\xx + b_g)$ for LLM fine-tuning, where $\ww_g$ and $b_g$ are learnable parameters. This gate acts as a soft switch, activating (values near~$1$) for fine-tuning inputs and suppressing (values near~$0$) for pre-training inputs. To further gain expressiveness, we extend this scalar gate to a vector-valued one, assigning a separate gate to each rank-one component of the modification. This yields a gating vector $\sigma(\WW_g \xx + \bb_g) \in [0,1]^r$, which we formalize in the next section.

\section{\ours{}: Dynamic Input-Sensitive \lora{}}\label{sec:method}

In Sec.~\ref{sec:background}, we saw that applying a fixed correction $\bDelta$ for all inputs induces a compromise between task adaptation and preserving the pre-trained mapping, whereas the Bayes-optimal predictor avoids this by conditioning the correction on the input. Thus, let us consider a general adapter of the form
\begin{equation}\label{eq:adapted}
    \yy = \WW_0 \xx + f(\xx; \btheta),
\end{equation}
where $f : \R^{d_x} \to \R^{d_y}$ is a trainable mapping. The desired behavior for $f$ is to provide the necessary adaptation on inputs from $p_{\text{ft}}$, while vanishing on inputs from $p_{\text{pt}}$. Meanwhile, the method should remain memory and computationally efficient to ensure practical applicability.

\textbf{Method.}
To this end,  
we build our approach based on \lora{}, a widely adopted PEFT method, which freezes $\WW_0$ and learns a low-rank correction as a product $\AA\BB\xx$ with $\AA \in \R^{d_y \times r}$ and $\BB \in \R^{r \times d_x}$. The key idea behind \ours{}
is to multiply each rank-one component by an input-dependent scalar gate $g_i(\xx) \in [0,1]$.
Collecting these into a gate vector $g(\xx) = (g_1(\xx),\dots,g_r(\xx))^\top$ and a diagonal matrix
$\GG(\xx) = \operatorname{diag}(g(\xx))$, the \ours{} adaptation is defined as
\begin{equation*}\label{eq:ours}
    f(\xx;\btheta) = \AA \GG(\xx) \BB \xx = \sum_{i=1}^{r} g_i(\xx) \aa_i \bb_i^\top\xx,
    \tag{\ours}
\end{equation*}
where $\aa_i$ and $\bb_i^\top$ are $\AA$'s columns and $\BB$'s rows, respectively. The gate vector is parameterized as
\begin{equation}\label{eq:gate}
    g(\xx) = \sigma \big( \WW_g \xx + \bb_g \big) \in [0,1]^r,
\end{equation}
with $\WW_g \in \R^{r \times d_x}$, $\bb_g \in \R^r$, and $\sigma(z) = 1/(1+e^{-z})$ applied element-wise. We use a sigmoid rather than a softmax so that gates act as independent soft switches, allowing multiple rank-one components to be active at the same time, or all inactive simultaneously, for the same input.

\textbf{Memory and inference overhead.}
\ours{} introduces additional $r d_x + r$ parameters per adapted block, with a total of $r d_y + 2 r d_x + r$ trainable parameters, comparable to $r d_y + r d_x$ of \lora{}. The gate computation requires $\mathcal{O}(d_x r)$ arithmetic operations, negligible relative to the cost of the linear projection $\WW_0 \xx$. Overall, both the memory and computational overhead are comparable to \lora{}.

\textbf{Initialization and approximation of the Bayes predictor.}
During fine-tuning, we typically observe only data from $p_{\text{ft}}$, whereas the Bayes-optimal predictor relies on both populations to identify which inputs require adaptation. Without access to $p_{\text{pt}}$, we cannot directly enforce $f(\xx;\btheta)\approx\zeroVec$ on pre-training-domain inputs; instead, we encode this as the \emph{default} behavior at initialization. Specifically, we initialize $\bb_g$ to a small negative value, such that $\sigma(\bb_g)$ is approximately zero, and initialize $\WW_g$ using a standard scheme, \emph{e.g.}, \citet{he2015delving}. Combined with the standard \lora{} initialization, this makes \ours{} start equal to the pre-trained model on all inputs. During training, gates are activated only when this reduces the FT loss, yielding an approximation to the Bayes-optimal input-dependent correction. The initialization of $\bb_g$ is a hyperparameter that may depend on the task and dataset; in our \llama{} and \mistral{} experiments, we set $\bb_g = -3\cdot\mathbf{1}$, where $\sigma(-3) \approx 0.05$.

\textbf{Gates learning vs. $\AA,\BB$ fine-tuning.}
The factors $\AA$ and $\BB$ play the same role as in standard \lora{}: they parameterize the adaptation itself and thus are tuned with a small learning rate. In contrast, the gate parameters $(\WW_g, \bb_g)$ learn a new and different task: to distinguish fine-tuning inputs from the rest of the input space, so that the update $\AA\BB\xx$ is applied selectively. This is more akin to representation learning during pre-training than to a small refinement of an existing linear map. Accordingly, we treat $(\AA, \BB)$ and $(\WW_g, \bb_g)$ as two distinct optimization problems and assign them \emph{separate learning rates}, with the gate learning rate substantially larger. In our \llama{} and \mistral{} experiments, we use a learning rate $5\times$ larger for the gate parameters than for $\AA$ and $\BB$.

\section{Experiments}\label{sec:experiments}
In this section, we evaluate the downstream performance of \ours{} in two settings: (i) fine-tuning \roberta{}-base~\cite{liu2019roberta} on the \textsc{GLUE} benchmark~\citep{wang2018glue} (Section~\ref{sec:exp-roberta}), and (ii) instruction tuning of \llama{}~$2$-$7$B~\cite{touvron2023llama} and \mistral{}-$7$B~\cite{jiang2023mistral7b} on mathematical reasoning and code generation tasks (Section~\ref{sec:exp-llm}). For each setting, we report two quantities: fine-tuning task performance and forgetting on benchmarks outside the fine-tuning domain. We compare \ours{} against full fine-tuning and three widely used PEFT baselines: \lora{}~\citep{hu2022lora}, \textsc{DoRA}~\citep{liu2024dora}, and \textsc{AdaLoRA}~\citep{zhang2023adalora}.

\begin{figure}[t]%
    \includegraphics[width=\linewidth]{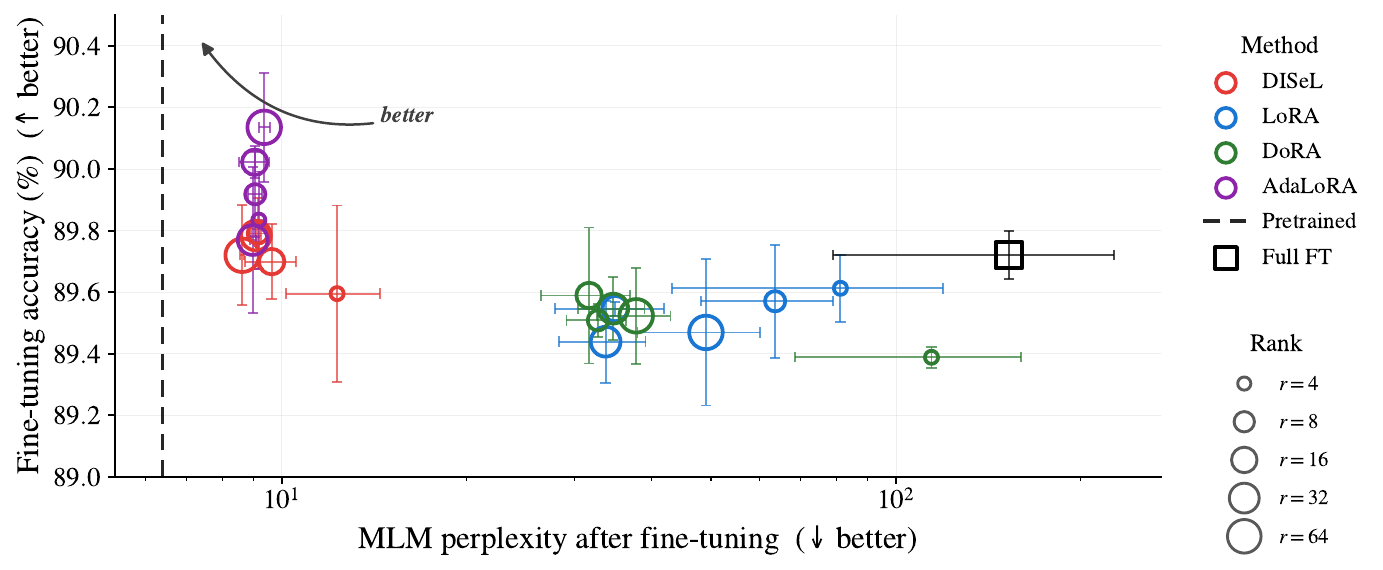}%
    \caption{{\bf \roberta{} on \textsc{Glue}.} We fine-tune \roberta{}-base on five \textsc{Glue} tasks: \textsc{MNLI}, \textsc{SST}-$2$, \textsc{QNLI}, \textsc{CoLA}, and \textsc{MRPC}, using various methods and ranks. The vertical axis measures the average test accuracy for the corresponding tasks, while the horizontal axis indicates the average masked-LM perplexity, evaluated on \textsc{BookCorpusOpen}, \textsc{CC-News}, and \textsc{WikiText}-$2$. Each point represents a method-rank pair averaged across three seeds, with error bars indicating the standard deviation. The dashed line marks the pre-trained baseline. Here, \ours{} and \textsc{AdaLoRA} remain close to this baseline while matching the FT accuracy of full FT, whereas the other methods forget much more.}
    \label{fig:roberta} 
\end{figure}

\subsection{\roberta{} on \textsc{Glue}}\label{sec:exp-roberta}
\textbf{Setup.} We fine-tune \roberta{}-base on five \textsc{GLUE} tasks: \textsc{MNLI}, \textsc{SST-2}, \textsc{QNLI}, \textsc{CoLA}, and \textsc{MRPC}. For each task, we consider ranks $r \in \{4, 8, 16, 32, 64\}$. Fine-tuning performance is measured as the average test accuracy across tasks. The forgetting is quantified via masked language modeling (MLM) perplexity of the fine-tuned model on \textsc{BookCorpusOpen}~\citep{bandy2021addressing}, \textsc{CC-News}~\citep{mackenzie2020cc}, and \textsc{WikiText}-$2$~\citep{merity2016pointer}, where the pre-trained model achieves an averaged perplexity of $\approx 6.4$ on these datasets. Each configuration (method, task, rank) is evaluated over three random seeds, and we report the mean $\pm$ standard deviation. Full details are provided in Appendix~\ref{sec:appendix-roberta}.

Figure~\ref{fig:roberta} plots the FT accuracy against retention. While most methods achieve comparable FT accuracy, they differ substantially in retention performance. Full fine-tuning attains high accuracy but deviates most strongly from the pre-trained baseline, yielding a mean perplexity of roughly $150$. Fixed-rank adapters such as \lora{} and \textsc{DoRA} mitigate this effect at the cost of slightly reduced FT accuracy, but still incur substantial forgetting, with perplexity ranging from $30$ to $100$. By contrast, \ours{} and \textsc{AdaLoRA} remain close to the pre-trained baseline, with perplexity around $9$, while matching or surpassing the accuracy of full fine-tuning. \ours{} thus achieves a similarly favorable tradeoff between accuracy and forgetting as pruning-based \textsc{AdaLoRA}, without requiring a pruning schedule. Although \textsc{AdaLoRA} performs competitively on this benchmark, the next section shows that its FT accuracy degrades at larger scales and on more challenging tasks, such as mathematical reasoning and code generation.

\subsection{\llama{} and \mistral{} on mathematical reasoning and code generation}\label{sec:exp-llm}
\textbf{Setup.} We extend our evaluation to large-scale instruction fine-tuning tasks. Specifically, we fine-tune \llama{}~$2$-$7$B on \textsc{MetaMath}~\citep{yu2024metamath} for math reasoning, and \mistral{}-$7$B on \textsc{Magicoder}~\citep{wei2023magicoder} for code generation. We evaluate FT performance on \textsc{GSM8K}~\citep{cobbe2021training} for math, and \textsc{HumanEval}~\citep{chen2021evaluating} for code. For each model, we use ranks $r \in \{16, 32, 64, 128, 256\}$. To measure forgetting, we evaluate retention on $14$ out-of-domain benchmarks, spanning commonsense reasoning, world knowledge, reading comprehension, and medical question answering, and report their unweighted mean accuracy. Appendix~\ref{sec:appendix-llm} lists the full benchmark suite and per-method hyperparameters.

\begin{figure}[t]%
    \begin{subfigure}[t]{0.5112\linewidth}%
        \includegraphics[width=\linewidth]{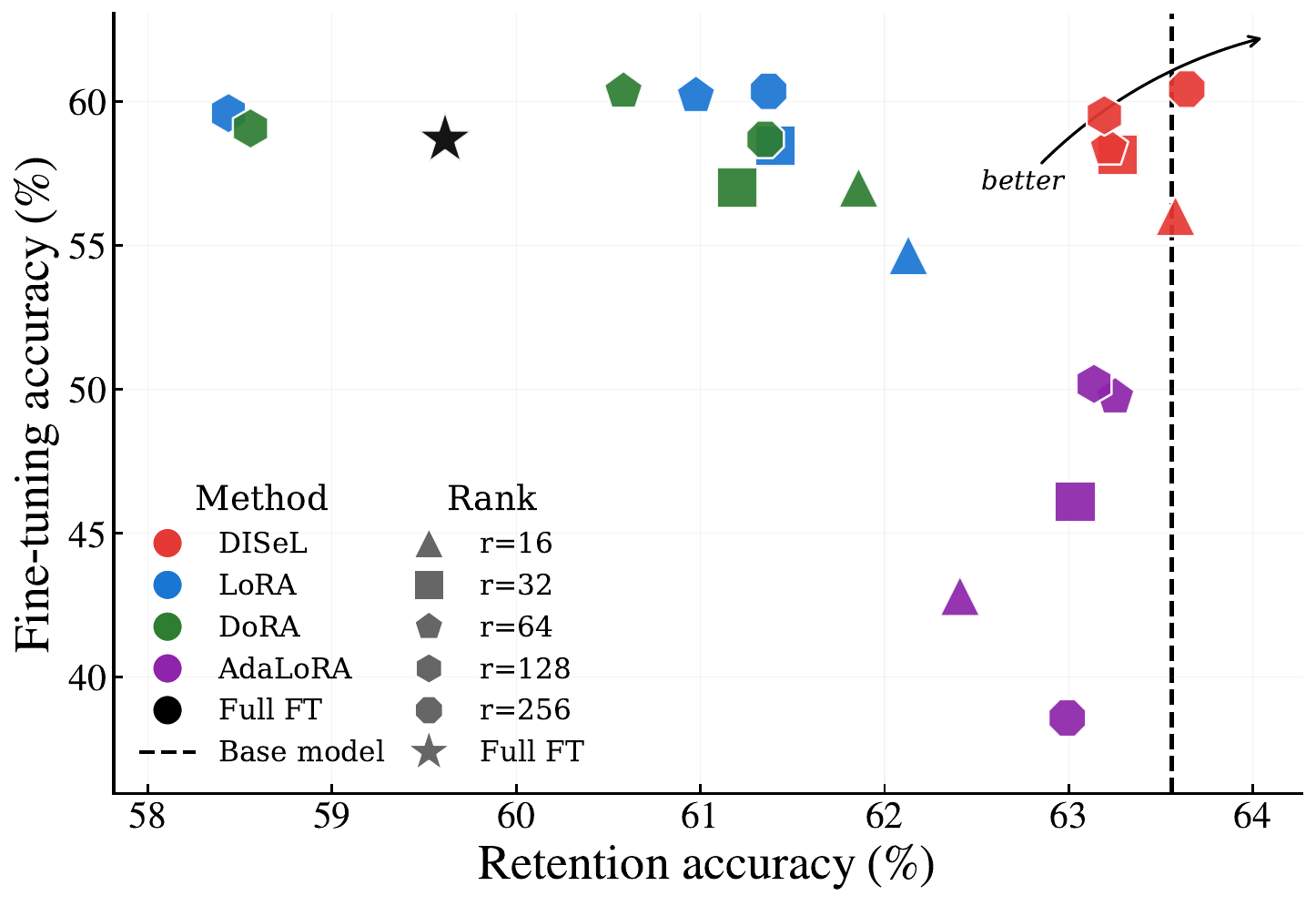}
        \caption{\llama{} math fine-tuning}
        \label{fig:llama-tradeoff}
    \end{subfigure}%
    \begin{subfigure}[t]{0.4888\linewidth}%
        \includegraphics[width=\linewidth, trim={1.1cm 0 0 0}, clip]{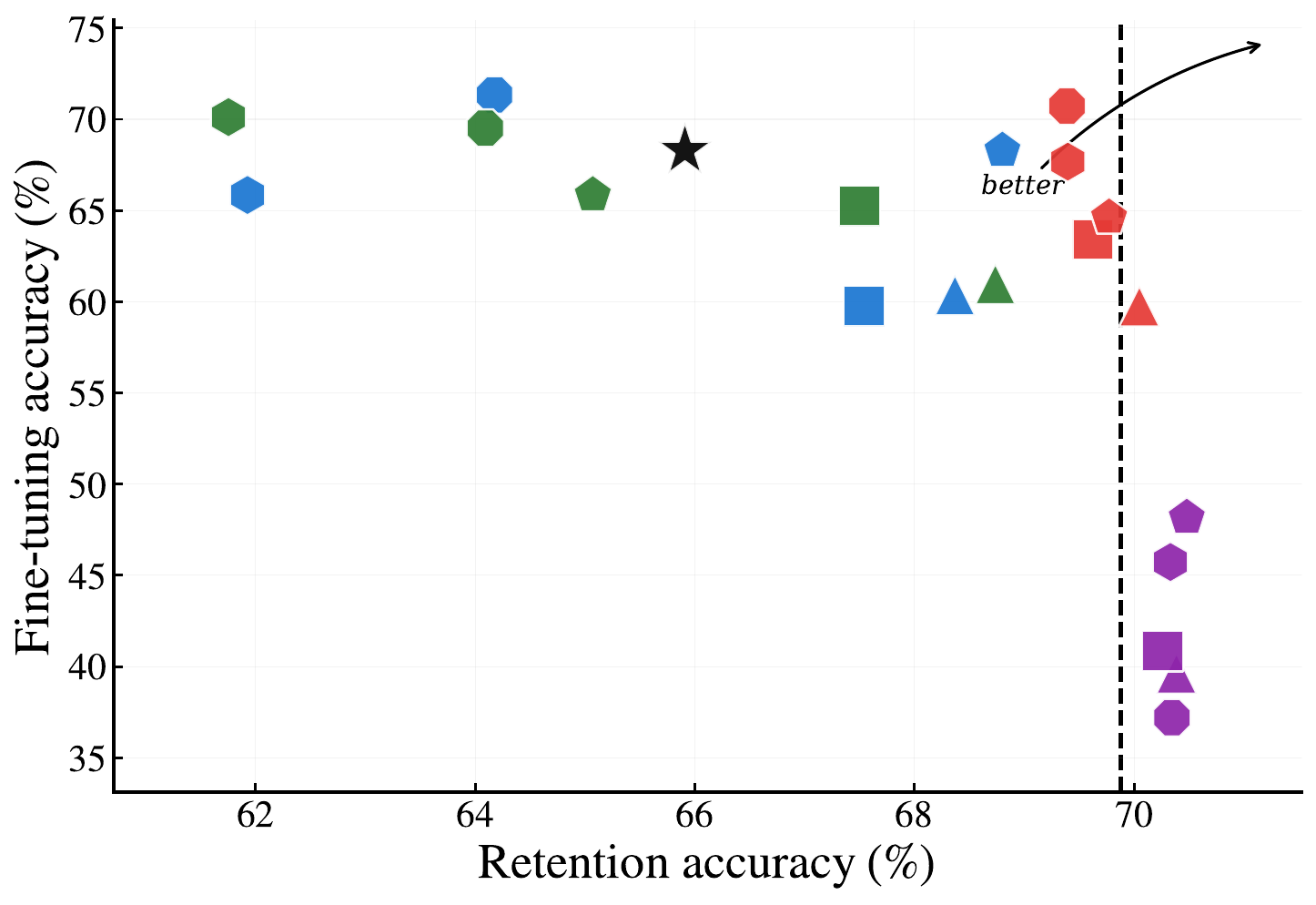}
        \caption{\mistral{} code fine-tuning}
    \label{fig:mistral-tradeoff}
    \end{subfigure}
    \caption{\textbf{\llama{} and \mistral{} instruction fine-tuning.}
    We fine-tune \llama{} and \mistral{} on math and code instructions, respectively, with various methods using different ranks. Panel \protect\subref{fig:llama-tradeoff} shows the results for \llama{}, whereas Panel \protect\subref{fig:mistral-tradeoff} depicts them for \mistral{}. Here, the vertical axis shows test accuracy for the corresponding fine-tuning task, and the horizontal axis shows the average accuracy across several pre-training tasks. The dashed vertical line marks the accuracy of the pre-trained model. This experiment shows that \ours{} achieves competitive fine-tuning accuracy without compromising performance on the pre-training tasks.}
    \label{fig:llm-tradeoff}
\end{figure}

Figure~\ref{fig:llm-tradeoff} visualizes the tradeoff between fine-tuning and retention. Here, \ours{} achieves strong task accuracy while maintaining retention close to the base model. On \llama{}, it reaches around $60\%$ FT accuracy with retention near $63.3\%$, only slightly below the base model. On \mistral{}, our approach exceeds $70\%$ FT accuracy while maintaining retention within a fraction of a percentage point of the base model ($\approx 70\%$). The occasional improvement over base-model retention suggests that fine-tuning can also improve the general reasoning abilities probed by some retention benchmarks. 

In contrast, full FT attains reasonable task accuracy but incurs substantial forgetting, shifting far left of the base-model retention line and failing to dominate the adapter methods. Its relatively small forgetting compared with the worst adapter runs is partly explained by the smaller learning rate used for fine-tuning. Among the three other PEFT methods, \textsc{AdaLoRA} matches \ours{} in retention, as in the \roberta{} experiments, but its FT accuracy drops to roughly $40$--$50\%$ on both models. \lora{} and \textsc{DoRA} perform competitively at low ranks, but their retention deteriorates as rank increases, losing $4$--$8\%$ at the largest ranks. By comparison, \ours{} achieves consistent performance across ranks, models, and datasets, preserving both task performance and pre-training capability. 

This robustness is practically useful beyond the final accuracy--retention tradeoff itself. Recent work has shown that \lora{}'s performance can be highly sensitive to learning rate, rank, initialization, and scaling choices, often requiring careful tuning to achieve strong performance~\citep{chen2026learning}. In contrast, the gating mechanism of \ours{} provides implicit regularization by limiting the effective modification of inactive or weakly activated components, thereby stabilizing training at larger or suboptimal adapter learning rates. Moreover, the separate learning rate for the gates also provides a direct control knob for the plasticity--retention tradeoff: larger gate learning rates encourage more gates to open and favor adaptation to the new task, whereas smaller gate learning rates keep more directions closed and favor retention of pre-training knowledge.

\section{Interpretability of \ours{}}\label{sec:interpretability}

\textbf{Setup.}
The main idea behind \ours{} is that its gates are \emph{input-dependent}. Specifically, the adapter correction should be active for inputs from the fine-tuning task, and suppressed for others. Here, we examine whether this behavior indeed emerges in trained models by analyzing the gate activations during inference. For this, we use held-out sequences from three domains: math, general text, and code. Math samples are drawn from the \textsc{GSM8K} test set, formatted with the same instruction--response template used during \llama{} fine-tuning. General-text samples are raw passages from the \textsc{WikiText}-$2$ test set. Code samples are taken from \textsc{HumanEval} test problems. For math and code, we consider gates on the response tokens, whereas for raw general-text, we measure gates over all tokens. For each sample, we record the post-sigmoid values at each adapted layer during the forward pass. We then aggregate the gate values, dividing the layers into three depth groups (early, mid, late), and create a normalized histogram for each group. Throughout this section, we present results for the rank-$32$ \llama{}~$2$-$7$B adapter fine-tuned for mathematical reasoning from Section~\ref{sec:exp-llm}. Additional results for \llama{} and \mistral{} are given in Appendix~\ref{sec:appendix-interpretability}.

\begin{figure}[t]%
    \includegraphics[width=\linewidth]{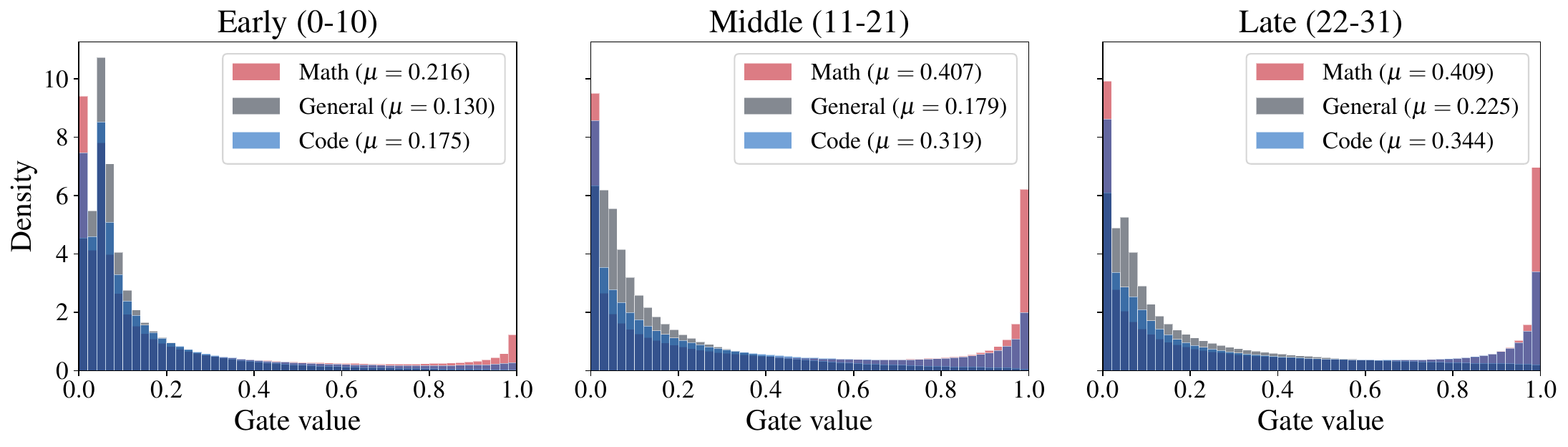}
    \caption{\textbf{Gate activation histograms across domains.}
    We analyze gate activations of a rank-$32$ \llama{} math adapter on held-out samples from math, general text, and code. Each panel corresponds to a layer-depth band: early, middle, or late layers. Here, the gates are substantially more open for math inputs, the fine-tuning domain, than for general text from the pre-training distribution. The largest differences appear in the middle and later layers. This demonstrates that \ours{}  selectively applies adaptation on inputs from the fine-tuned distribution.}
    \label{fig:llama-gate-domain}
\end{figure}

\textbf{Domain analysis.}
Figure~\ref{fig:llama-gate-domain} shows gate activations across different input domains. In early layers, the gates are nearly closed across all domains, suggesting that adaptation in these layers might not be needed. Yet, in the middle and late layers, there is a clear separation: math samples have substantially larger gate values than general text, while code lies between the two. This ordering makes sense in this setup: math is the fine-tuning domain, general text is outside that domain, and code shares some structured-symbolic and logical reasoning with math. Overall, the gates are not static; they open more to inputs from the fine-tuning domain than to those from the pre-trained domain.

\begin{figure}[t]%
    \includegraphics[width=\linewidth]{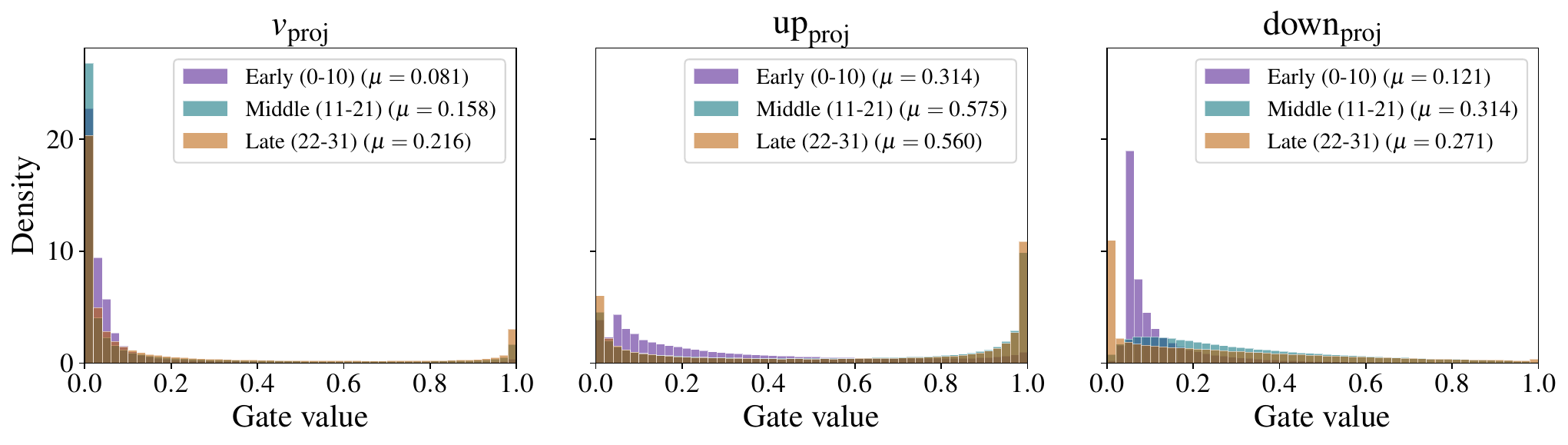}
    \caption{\textbf{Layer-wise gate usage within key modules.}
    We processed held-out math samples with \llama{} using a rank-$32$ \ours{} adapter and measured the output distributions of each module's gates. This figure presents these distributions for the attention value projection ($v_{\mathrm{proj}}$), the MLP up projection ($\mathrm{up}_{\mathrm{proj}}$), and the MLP down projection ($\mathrm{down}_{\mathrm{proj}}$). Each panel shows normalized histograms for early, middle, and late layers. In this example, the gates are most open in the MLP up projection, especially in the middle and late layers, while the value and down projections exhibit more moderate activation. This demonstrates that \ours{} learns not only when to activate adapter components, but also where in the model those components are most useful.}
    \label{fig:llama-gate-modules}
\end{figure}

\textbf{Module analysis.}
In Figure~\ref{fig:llama-gate-modules}, we fix the input domain to held-out math examples and compare selected adapted layer types: the attention module ($v_{\mathrm{proj}}$), the MLP up projection ($\mathrm{up}_{\mathrm{proj}}$), and the MLP down projection ($\mathrm{down}_{\mathrm{proj}}$). Here we see that gate activation is highly non-uniform across different module types. The MLP up projection has the largest gate values, especially in middle and late layers; the down projection shows a weaker version of the same trend; and the value projection remains comparatively closed. This suggests that the learned gates provide a useful diagnostic for where task-specific adapter capacity is used, and may help improve the design of future modules. Taken together, Figures~\ref{fig:llama-gate-domain} and~\ref{fig:llama-gate-modules} demonstrate that \ours{} learns a conditional routing pattern: gate usage varies across input domains and concentrates in specific parts of the network rather than being distributed uniformly across all ranks, layers, and modules. 

\section{Retention over time}\label{sec:retention_over_time}

The tradeoff plots in Section~\ref{sec:experiments} report retention at the best target-task checkpoint, but do not reveal how retention \emph{evolves} during training. A method may end close to the base model while passing through degraded intermediate states, or it may lose capability steadily from the first few optimization steps. To distinguish these regimes, we track retention at every saved checkpoint of the \llama{}~$2$-$7$B, for ranks $r \in \{16, 32, 64, 128, 256\}$. As before, retention is measured as the mean accuracy over the $14$-benchmark suite from Section~\ref{sec:exp-llm}. Additional results are provided in Appendix~\ref{sec:appendix-retention-over-time}.

Figure~\ref{fig:retention-over-time} shows the resulting trajectories for \ours{} (left) and \lora{} (right). For \lora{}, retention degrades progressively with rank: at $r{=}16$, it remains within about one percentage point of the base model, but at higher ranks it exhibits a clear downward drift, falling roughly a $5\%$ at the largest ranks. This degradation is persistent rather than transient; once lost, performance does not recover within the training horizon. In contrast, \ours{} maintains near-flat retention curves across all ranks, with deviations remaining within about half a percentage point of the base model from the first checkpoint to the last. This stability can be attributed to the fact that input-dependent gates keep the correction largely inactive on inputs outside the fine-tuning distribution, preventing increased capacity from inducing additional drift. As a result, \ours{} effectively decouples adapter rank from forgetting, enabling the use of larger ranks without sacrificing retention.

Finally, we observe that forgetting could also depend on the optimization setting. For \lora{}, the $r{=}128$ run exhibits more forgetting than $r{=}256$, as the latter required a smaller learning rate ($10^{-4}$) for stability, whereas the former uses a twice larger step size. A similar effect appears in Figure~\ref{fig:llm-tradeoff}, where Full FT does not incur the largest forgetting, partly due to its smaller learning rate used ($10^{-5}$).

\section{Related work}\label{sec:related}
We discuss three lines of work most related to \ours{}: parameter-efficient low-rank fine-tuning, input-conditional and mixture-style adapters, and continual learning of pre-trained models.

\textbf{Parameter-efficient and low-rank fine-tuning.}
\lora{} \citep{hu2022lora} approximates the FT update using a fixed low-rank correction $\bDelta = \AA\BB$ added to a frozen pre-trained weight, and has become the dominant approach to parameter-efficient adaptation. A large body of follow-up work refines the \emph{parameterization} of~$\bDelta$: \textsc{AdaLoRA}~\citep{zhang2023adalora} adapts the rank budget across layers, \textsc{DoRA}~\citep{liu2024dora} decouples magnitude from direction, \textsc{PiSSA}~\citep{meng2024pissa} initializes $\AA, \BB$ from the principal singular subspace of~$\WW$, \textsc{LoRA-GA}~\citep{wang2024lora} chooses the initialization such that the first gradient step approximates full FT, \textsc{VeRA}~\citep{kopiczko2024vera} freezes shared random $\AA, \BB$ and learns only a diagonal scaling between them, structurally close to the $\GG(\xx)$ in \ours{} but fixed across inputs, and \textsc{AuroRA}~\citep{dong2026aurora} introduces a nonlinear coupling between the factors. Other variants modify optimization dynamics: \textsc{LoRA+}~\citep{hayou2024lora+} uses asymmetric learning rates for $\AA$ and $\BB$, \textsc{LoRAM}~\citep{loram} optimizes the low-rank adapter on Riemannian manifolds, and \textsc{QLoRA}~\citep{dettmers2023qlora} combines low-rank adaptation with weight quantization. Despite their differences, these methods produce a single correction matrix that is uniformly applied to every input during inference. \ours{} preserves the low-rank parameterization and matches the computational efficiency of \lora, while making the correction input-dependent through lightweight gates. This selective adaptation mechanism directly addresses the structural conflict identified in Section~\ref{sec:background}.

\textbf{Input-conditional and mixture-of-\lora{} approaches.}
A separate line of work introduces input- or task-conditional adaptation by combining \emph{multiple} \lora{} modules. \textsc{MoLE}~\citep{wu2024mixture} routes between several \lora{} adapters via a learned gating network, in the spirit of mixture-of-experts. \textsc{LoraHub}~\citep{huang2023lorahub} composes pre-trained task-specific \lora{} modules into a single weighted combination selected for each downstream task. These approaches operate at the coarse-grained level of entire adapters, and scale both the parameter count and the routing cost with the number of modules. The closest architectural prior is \textsc{Gated~LoRA}~\citep{eom2025gatedlora}, which treats each rank-one component as a mini-expert, and selects the active components based on the input. Specifically, it uses a ReLU activation function on the same down-projection features used to compute the \lora{} update, thereby avoiding a separate router. Its primary goal is to improve parameter utilization and reduce inter-task interference, especially in multi-task fine-tuning. Similar to \textsc{Gated~LoRA}, \ours{} also gates rank-one components inside a single adapter. However, \ours{} uses separate learnable sigmoid gates initialized near zero so that the default behavior matches the pre-trained model. More importantly, our objective is fundamentally different: rather than improving multi-task routing, \ours{} is designed to preserve pre-trained behavior on inputs outside the fine-tuning distribution.

\textbf{Continual learning.}
Continual learning methods adapt a pre-trained model to a sequence of tasks while preserving previously acquired knowledge. \textsc{GainLoRA}~\citep{liang2025gated} introduces a separate \lora{} branch for each new task and learns gates that limit interference between branches. Related prompt-based methods, including \textsc{DualPrompt}~\citep{wang2022dualprompt}, \textsc{S-Prompts}~\citep{wang2022s}, \textsc{HiDe-Prompt}~\citep{wang2023hierarchical}, and mixture-style prefix tuning~\citep{le2024mixture}, route inputs among task- or domain-specific prompts, while \textsc{RanPAC}~\citep{mcdonnell2023ranpac} avoids backbone updates entirely by learning class prototypes after a fixed random projection. These methods address forgetting across a sequence of tasks and typically operate at the level of entire adapters, prompts, or task-specific classifiers. \ours{} instead targets the retention--adaptation tradeoff within a \emph{single} fine-tuning task, using continuous input-dependent gating over rank-one components inside a single \lora{} adapter, without requiring task identities or routing labels.

\begin{figure}[t]%
    \begin{subfigure}[t]{0.5112\linewidth}%
        \includegraphics[width=\linewidth]{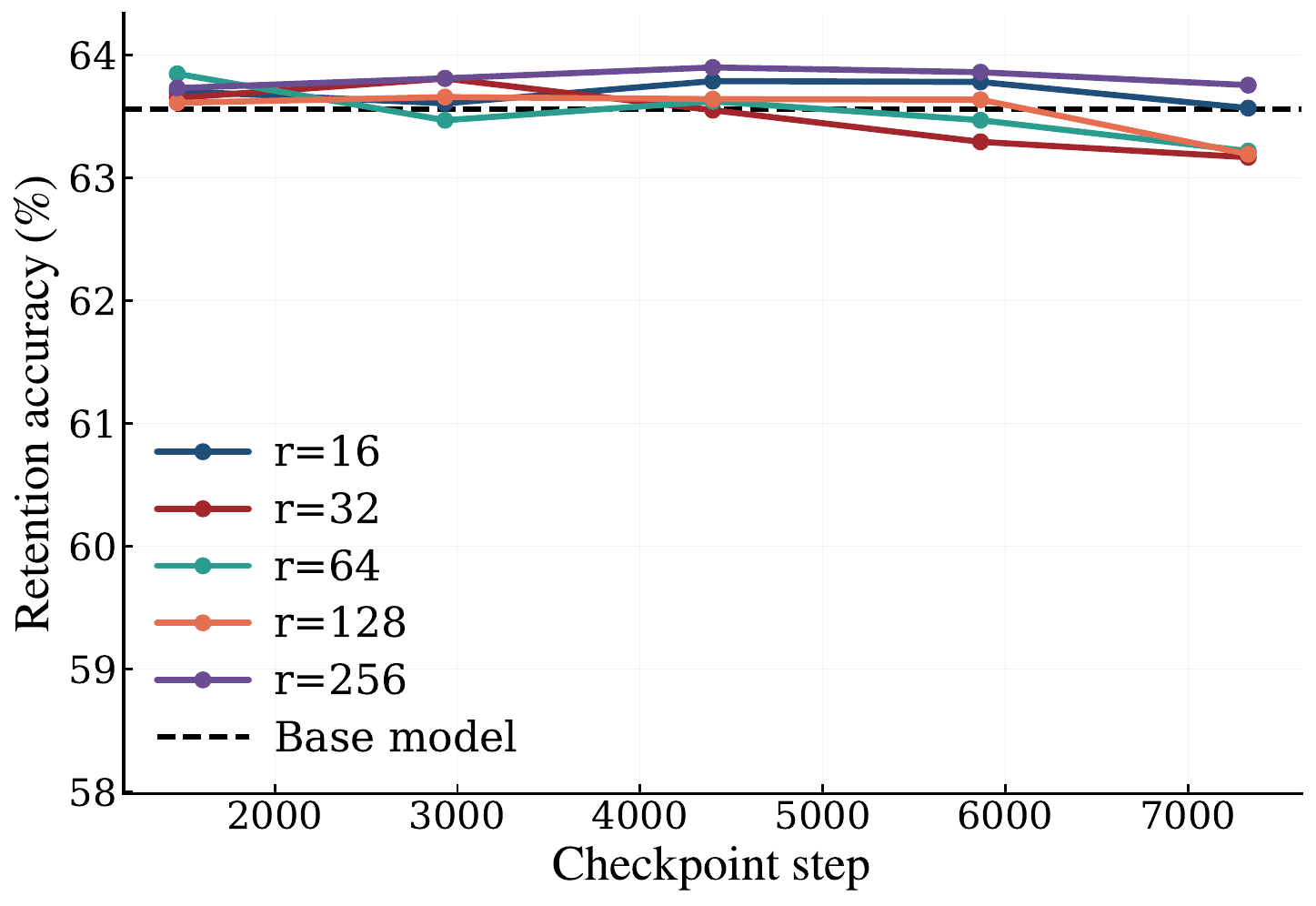}
        \caption{\ours{}}%
        \label{fig:llama-retention-dial}%
    \end{subfigure}%
    \begin{subfigure}[t]{0.4888\linewidth}%
        \includegraphics[width=\linewidth, trim={1.1cm 0 0 0}, clip]{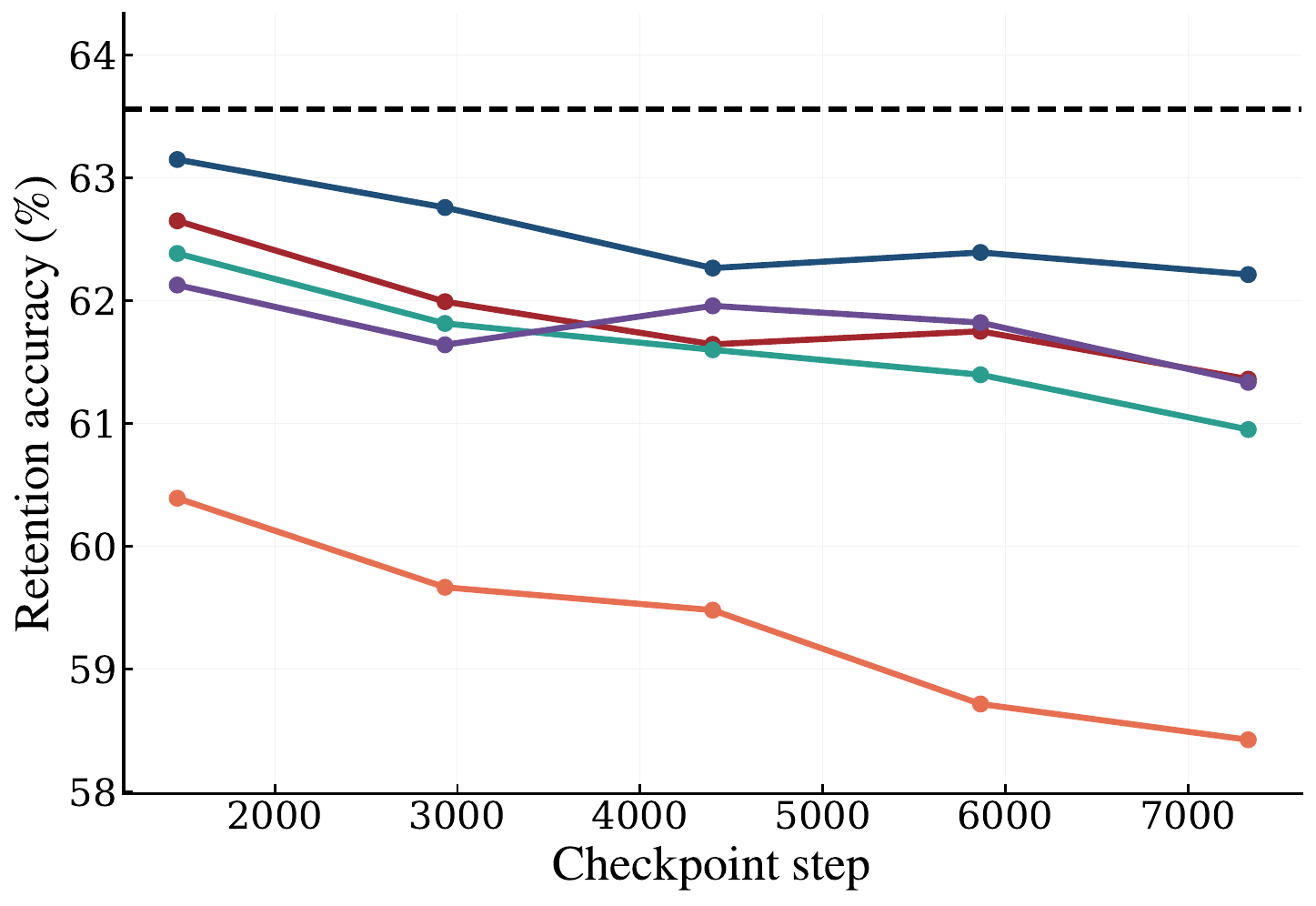}%
        \caption{\lora{}}%
        \label{fig:llama-retention-lora}%
    \end{subfigure}%
    \caption{\textbf{Retention vs. training step.} We evaluate the retention performance during fine-tuning from the experiment in Sec.~\ref{sec:exp-llm}. Here we present the results for \llama{}~$2$-  $7$B model fine-tuned on \textsc{MetaMathQA} using \lora{} and \ours{} with various ranks. Panel~\protect\subref{fig:llama-retention-dial} shows the result for \ours{}, while Panel~\protect\subref{fig:llama-retention-lora} plots them for \lora{}. We see that \ours{} maintains near-zero retention drop at every rank, including the largest, whereas \lora{}'s retention deteriorates with both rank and training time.}
    \label{fig:retention-over-time}
\end{figure}

\section{Conclusion}\label{sec:conclusion}
We introduced \ours{}, an FT method that addresses the input-agnostic nature of standard adapter updates. While \lora{} learns a fixed correction that is applied to every input, \ours{} makes this correction input-dependent by placing lightweight gates on individual rank-one components. This design is motivated by the tension identified in our linear analysis: a fixed update must compromise between adapting to fine-tuning-domain inputs and preserving the pre-trained mapping elsewhere.

Across \roberta{} on GLUE and instruction tuning of \llama{} and \mistral{} models on mathematical reasoning and code generation, \ours{} achieves a favorable tradeoff between FT accuracy and retention. It keeps retention close to the base model while maintaining competitive task performance. The learned gates further strengthen the interpretability angle of \ours{}: they reveal when the adapter is activated, which inputs induce stronger adaptation, and where in the network task-specific capacity is concentrated. This makes \ours{} useful not only as a retention-preserving FT method, but also as a diagnostic tool for understanding and potentially controlling adaptation.

\textbf{Limitations and future work.} \ours{} does not guarantee preservation of all pre-trained capabilities, and its behavior depends on choices such as gate initialization, gate learning rate, and the adapted modules. Moreover, while effective in single-task FT, it remains unclear how best to extend input-dependent adaptation to multi-task or continual learning, or how to merge the learned adapter into the base weights for sustainable deployment. Future work could combine \ours{} with explicit retention objectives, use gate statistics for adaptive rank selection, and evaluate the method in broader settings such as alignment, multilingual transfer, long-context adaptation, and continual learning.

\section*{Acknowledgment}
This work was supported by the Helmholtz Association's Initiative and Networking Fund on the HAICORE@FZJ partition. Part of this work was carried out while S.U.\ Stich was visiting the Simons Institute for the Theory of Computing. We also gratefully acknowledge funding from the European Research Council (ERC) under the Horizon Europe Framework Programme (HORIZON) for proposal number 101170430 CollectiveMinds. Views and opinions expressed are however those of the authors only and do not necessarily reflect those of the European Union or the European Research Council. Neither the European Union nor the granting authority can be held responsible for them.

\bibliographystyle{plainnat}
\bibliography{references}


\clearpage
\appendix

\begin{center}
{\Huge\bfseries Appendix}
\end{center}
\renewcommand{\thesection}{\Roman{section}}

\section{Complete details of the motivating example}\label{sec:appendix-toy}

This appendix gives the full derivation for the linear-regression example in Section~\ref{sec:background}. The purpose of the toy example is to isolate the structural limitation of a fixed, input-independent correction: the same update is applied both to inputs requiring adaptation and to inputs whose pretrained behavior should be preserved

\paragraph{Setup.}
Let $\WW_0 \in \R^{d\times d}$ be the frozen pre-trained linear map. Inputs are sampled from the symmetric mixture of a fine-tuning population $p_{\mathrm{ft}}$ and a pre-training population $p_{\mathrm{pt}}$:
\begin{equation}
    \xx \sim \begin{cases}
        p_{\mathrm{ft}} & \text{w.p. } \tfrac12,\\
        p_{\mathrm{pt}} & \text{w.p. } \tfrac12.
    \end{cases}
\end{equation}
The target is
\begin{equation}
    \yy =
    \begin{cases}
        (\WW_0+\MM)\xx & \text{when } \xx \sim p_{\mathrm{ft}},\\
        \WW_0\xx & \text{when } \xx \sim p_{\mathrm{pt}},
    \end{cases}
    \label{eq:appendix-toy-target}
\end{equation}
where $\MM\in\R^{d\times d}$ is the task-specific update. Thus the desired correction is $\MM\xx$ on fine-tuning inputs and $\zeroVec$ on pre-training inputs.

\subsection{The best fixed correction}
\label{subsec:appendix-fixed-correction}

Consider any fixed matrix correction $\bDelta\in\R^{d\times d}$. After subtracting the frozen term $\WW_0\xx$ from both the model and the target, the population loss is
\begin{equation}
    \mathcal{L}(\bDelta)
    =
    \tfrac12\,\E_{p_{\mathrm{ft}}}\!\left[\norm{(\bDelta-\MM)\xx}^2\right]
    +
    \tfrac12\,\E_{p_{\mathrm{pt}}}\!\left[\norm{\bDelta\xx}^2\right].
    \label{eq:appendix-fixed-loss}
\end{equation}
Let
\begin{equation}
    \CovMat^{\mathrm{ft}}_{\xx\xx} \coloneqq \E_{p_{\mathrm{ft}}}[\xx\xx^\top],
    \qquad
    \CovMat^{\mathrm{pt}}_{\xx\xx} \coloneqq \E_{p_{\mathrm{pt}}}[\xx\xx^\top]
\end{equation}
denote the uncentered second-moment matrices. Then
\begin{equation}
    \mathcal{L}(\bDelta)
    =
    \tfrac12\,\trace\!\left((\bDelta-\MM)\CovMat^{\mathrm{ft}}_{\xx\xx}(\bDelta-\MM)^\top\right)
    +
    \tfrac12\,\trace\!\left(\bDelta\CovMat^{\mathrm{pt}}_{\xx\xx}\bDelta^\top\right).
\end{equation}
Taking the derivative with respect to $\bDelta$ gives
\begin{equation}
    \nabla_{\bDelta}\mathcal{L}
    =
    (\bDelta-\MM)\CovMat^{\mathrm{ft}}_{\xx\xx}+\bDelta\CovMat^{\mathrm{pt}}_{\xx\xx}.
\end{equation}
Assuming $\CovMat^{\mathrm{ft}}_{\xx\xx}+\CovMat^{\mathrm{pt}}_{\xx\xx}$ is invertible, the unconstrained fixed-correction minimizer satisfies
\begin{equation}
    \bDelta^\star_{\mathrm{uncon}}
    =
    \MM\CovMat^{\mathrm{ft}}_{\xx\xx}(\CovMat^{\mathrm{ft}}_{\xx\xx}+\CovMat^{\mathrm{pt}}_{\xx\xx})^{-1}.
    \label{eq:appendix-uncon-fixed}
\end{equation}

Suppose the two populations have the same uncentered second moment:
\begin{equation}
    \CovMat^{\mathrm{ft}}_{\xx\xx}=\CovMat^{\mathrm{pt}}_{\xx\xx}=:\CovMat_{\xx\xx}.
\end{equation}
Then we have:
\begin{equation}
    \bDelta^\star_{\mathrm{uncon}}=\tfrac12\MM.
    \label{eq:appendix-half-m}
\end{equation}
Substituting this back into~\eqref{eq:appendix-fixed-loss},
\begin{align}
    \mathcal{L}(\bDelta^\star_{\mathrm{uncon}})
    &=
    \tfrac12\,\E_{p_{\mathrm{ft}}}\!\left[\norm{-\tfrac12\MM\xx}^2\right]
    +
    \tfrac12\,\E_{p_{\mathrm{pt}}}\!\left[\norm{\tfrac12\MM\xx}^2\right] \nonumber\\
    &=
    \tfrac18\,\E_{p_{\mathrm{ft}}}\!\left[\norm{\MM\xx}^2\right]
    +
    \tfrac18\,\E_{p_{\mathrm{pt}}}\!\left[\norm{\MM\xx}^2\right] \nonumber\\
    &=
    \tfrac14\,\E_{\xx\sim\frac12 p_{\mathrm{ft}}+\frac12 p_{\mathrm{pt}}}
    \!\left[\norm{\MM\xx}^2\right]
    \\
    &=\tfrac14 \trace(\MM\CovMat_{\xx\xx}\MM^\top).
    \label{eq:appendix-fixed-opt-loss}
\end{align}

This shows the tradeoff: the best fixed correction applies only half of the desired update on $p_{\mathrm{ft}}$ and applies a nonzero half-update on $p_{\mathrm{pt}}$.

\subsection{The Bayes-optimal input-dependent correction}
\label{subsec:appendix-bayes-correction}

Now allow the correction to be any measurable function $f:\R^d\to\R^d$. The loss becomes
\begin{equation}
    \mathcal{L}(f)
    =
    \tfrac12\,\E_{p_{\mathrm{ft}}}\!\left[\norm{f(\xx)-\MM\xx}^2\right]
    +
    \tfrac12\,\E_{p_{\mathrm{pt}}}\!\left[\norm{f(\xx)}^2\right].
    \label{eq:appendix-function-loss}
\end{equation}
Writing the expectations as integrals, the contribution of a fixed input $\xx$ is
\begin{equation}
    J_{\xx}(v)
    =
    \tfrac12\,p_{\mathrm{ft}}(\xx)\norm{v-\MM\xx}^2
    +
    \tfrac12\,p_{\mathrm{pt}}(\xx)\norm{v}^2,
\end{equation}
where $v=f(\xx)$. Since $J_{\xx}$ depends on $f$ only through its value at $\xx$, the optimal function is obtained by minimizing this quadratic separately for each $\xx$.

The first-order condition is
\begin{equation}
    p_{\mathrm{ft}}(\xx)(v-\MM\xx)+p_{\mathrm{pt}}(\xx)v=0,
\end{equation}
so the Bayes-optimal correction is
\begin{equation}
    f^\star(\xx)
    =
    \pi_{\mathrm{ft}}(\xx)\MM\xx,
    \qquad
    \pi_{\mathrm{ft}}(\xx)
    \coloneqq
    \frac{p_{\mathrm{ft}}(\xx)}{p_{\mathrm{ft}}(\xx)+p_{\mathrm{pt}}(\xx)}.
    \label{eq:appendix-bayes-predictor}
\end{equation}
Here $\pi_{\mathrm{ft}}(\xx)$ is the posterior probability that $\xx$ came from the fine-tuning population under the symmetric mixture. The corresponding Bayes loss is
\begin{equation}
    \mathcal{L}_{\mathrm{Bayes}}
    =
    \tfrac12\int
    \frac{p_{\mathrm{ft}}(\xx)p_{\mathrm{pt}}(\xx)}
    {p_{\mathrm{ft}}(\xx)+p_{\mathrm{pt}}(\xx)}
    \norm{\MM\xx}^2\,d\xx.
    \label{eq:appendix-bayes-loss}
\end{equation}
No input-dependent method can achieve a smaller squared-error loss on this population problem, because~\eqref{eq:appendix-bayes-predictor} is the pointwise minimizer of the loss integrand.

\subsection{\ours{} realizes the Bayes predictor}
\label{subsec:appendix-dial-realizes-bayes}
Assume that the two populations are multivariate Gaussian with the same covariance matrix but different means:
\begin{equation}
    p_{\text{ft}}(\xx) = \mathcal{N}(\xx;\,\bmu_{\mathrm{ft}},\,\CovMat),
    \qquad
    p_{\text{pt}}(\xx) = \mathcal{N}(\xx;\,\bmu_{\mathrm{pt}},\,\CovMat).
    \label{eq:appendix-shared-cov}
\end{equation}
Then we can compute the Bayes-optimal correction exactly as follows:
\begin{align}
    \log \frac{p_{\text{ft}}(\xx)}{p_{\text{pt}}(\xx)}
    &= -\tfrac12(\xx-\bmu_{\text{ft}})^\top\CovMat^{-1}(\xx-\bmu_{\text{ft}})
       + \tfrac12(\xx-\bmu_{\text{pt}})^\top\CovMat^{-1}(\xx-\bmu_{\text{pt}}) \nonumber \\
    &= (\bmu_{\text{ft}}-\bmu_{\text{pt}})^\top\CovMat^{-1}\xx
       \;-\; \tfrac12\!\left(\bmu_{\text{ft}}^\top\CovMat^{-1}\bmu_{\text{ft}}
    -\bmu_{\text{pt}}^\top\CovMat^{-1}\bmu_{\text{pt}}\right) \nonumber \\
    &\coloneqq \ww_g^\top \xx + b_g \;,
    \label{eq:appendix-log-ratio}
\end{align}
where 
\begin{equation}
    \ww_g \;\coloneqq\; \CovMat^{-1}(\bmu_{\text{ft}} - \bmu_{\text{pt}}) \;\in\; \R^d,
    \qquad
    b_g \;\coloneqq\; \tfrac12\!\left(
    \bmu_{\text{pt}}^\top\CovMat^{-1}\bmu_{\text{pt}} -\bmu_{\text{ft}}^\top\CovMat^{-1}\bmu_{\text{ft}}
    \right) \;\in\; \R.
    \label{eq:appendix-w-b}
\end{equation}
Substituting~\eqref{eq:appendix-log-ratio} into~\eqref{eq:appendix-bayes-predictor}, we have
\begin{equation}
    \pi_{\text{ft}}(\xx)
    \;=\; \frac{p_{\text{ft}}(\xx)}{p_{\text{ft}}(\xx)+p_{\text{pt}}(\xx)}
    \;=\; \frac{1}{1+\exp\bigl(-(\ww_g^\top \xx + b_g)\bigr)}
    \;=\; \sigma\!\rb{\ww_g^\top \xx + b_g},
    \label{eq:appendix-posterior-sigmoid-multivariate}
\end{equation}
and thus the Bayes-optimal correction is a sigmoid gate multiplying the task update:
\begin{equation}
    f^\star(\xx)
    =
    \sigma\!\rb{\ww_g^\top \xx + b_g} \MM\xx.
\end{equation}

\paragraph{Choice of \ours{} parameters that realize $f^\star$.}
Given any factorization $\MM = \AA\BB$ with $\AA \in \R^{d \times r}$ and $\BB \in \R^{r \times d}$ (which exists whenever $\rank(\MM) \le r$), set the gate parameters to
\begin{equation}
    \WW_g \;=\; \mathbf{1}_r\,\ww_g^\top \;\in\; \R^{r\times d},
    \qquad
    \bb_g \;=\; b_g\,\mathbf{1}_r \;\in\; \R^r,
    \label{eq:appendix-Wg-bg}
\end{equation}
where $\mathbf{1}_r = (1,\dots,1)^\top \in \R^r$ and $\ww_g, b_g$ are given by~\eqref{eq:appendix-w-b}. Each rank's gate then takes the same value
\begin{equation}
    g_i(\xx) \;=\; \sigma\!\rb{\ww_g^\top \xx + b_g} \;=\; \pi_{\mathrm{ft}}(\xx),
    \qquad i = 1, \dots, r,
\end{equation}
so $\GG(\xx) = \pi_{\mathrm{ft}}(\xx)\,\II_r$ and the \ours{} correction is
\begin{equation}
    \AA\,\GG(\xx)\,\BB\,\xx
    \;=\; \pi_{\mathrm{ft}}(\xx)\,\AA\BB\,\xx
    \;=\; \pi_{\mathrm{ft}}(\xx)\,\MM\,\xx
    \;=\; f^\star(\xx).
\end{equation}
Hence \ours{} represents the Bayes-optimal input-dependent correction exactly in this case. 

\subsection{Experimental setup for Figure~\ref{fig:toy}}
For the concrete experiment, 
we use $d=16$. 
$\MM=\UU\VV$ with $\UU\in\R^{d\times 2}$ and $\VV\in\R^{2\times d}$ sampled with i.i.d.\ standard Gaussian entries, so $\rank(\MM)=2$ almost surely. We write $\xx=(x_1,\xx_{2:d})$ and use
\begin{equation}
    p_{\mathrm{ft}}:\quad x_1\sim\mathcal{N}(+\mu,s^2),\quad \xx_{2:d}\sim\mathcal{N}(\zeroVec,\II_{d-1}),
\end{equation}
\begin{equation}
    p_{\mathrm{pt}}:\quad x_1\sim\mathcal{N}(-\mu,s^2),\quad \xx_{2:d}\sim\mathcal{N}(\zeroVec,\II_{d-1}),
\end{equation}
with $\mu=3$ and $s^2=0.25$. The two populations differ only in the sign of the mean of $x_1$.

Since the experiment uses \lora{} rank $r=2$ and $\rank(\MM)=2$, the matrix $\tfrac12\MM$ is feasible for rank-$2$ \lora{}. Hence, in this toy example, the rank-constrained \lora{} optimum is exactly the unconstrained fixed-correction optimum in~\eqref{eq:appendix-half-m}. The loss in~\eqref{eq:appendix-fixed-opt-loss} is therefore not only a lower bound; it is the actual best loss achievable by a fixed rank-$2$ correction.

\section{\roberta{} experiments: details and hyperparameters}\label{sec:appendix-roberta}

This appendix provides the full setup for the \roberta{}-base experiments on five  \textsc{Glue} tasks reported in Section~\ref{sec:exp-roberta} of the main paper.

\subsection{Model and tasks}

\paragraph{Architecture.} We use \roberta{}-base~\citep{liu2019roberta} ($125$M parameters) initialized from the HuggingFace \texttt{roberta-base} checkpoint. Each task uses a randomly initialized linear classification head trained jointly with the adapter.

\paragraph{Adapter target modules.} All adapters (and Full FT, when applicable) act on every linear layer in the encoder:
\begin{itemize}
    \setlength{\itemsep}{0pt}
    \item Attention: \texttt{q\_proj}, \texttt{k\_proj}, \texttt{v\_proj}, \texttt{attention.output.dense}.
    \item MLP: \texttt{intermediate.dense}, \texttt{output.dense}.
\end{itemize}

\paragraph{Tasks.} Five \textsc{Glue} tasks \citep{wang2018glue}: \textsc{MNLI}, \textsc{SST-2}, \textsc{QNLI}, \textsc{CoLA}, and \textsc{MRPC}. The underlying task datasets are \textsc{MultiNLI}, \textsc{SST}, \textsc{SQuAD}-derived \textsc{QNLI}, \textsc{CoLA}, and \textsc{MRPC} \citep{williams2018broad,socher2013recursive,rajpurkar2016squad,warstadt2019neural,dolan2005automatically}; sample counts and per-task validation metrics are listed in Table~\ref{tab:roberta-tasks}. Validation accuracy is reported per task; the headline score is their unweighted mean.

\begin{table}[ht]
\centering
\renewcommand{\arraystretch}{1.2}
\caption{The five \textsc{Glue} tasks used for fine-tuning \roberta{}-base.}
\label{tab:roberta-tasks}
\begin{tabular}{lrrl}
\toprule
\textbf{Task} & \textbf{\# train} & \textbf{\# classes} & \textbf{Validation metric} \\
\midrule
\textsc{MNLI}  & 392{,}702 & 3 & matched accuracy \\
\textsc{SST-2} &  67{,}349 & 2 & accuracy \\
\textsc{QNLI}  & 104{,}743 & 2 & accuracy \\
\textsc{CoLA}  &   8{,}551 & 2 & accuracy \\
\textsc{MRPC}  &   3{,}668 & 2 & accuracy \\
\bottomrule
\end{tabular}
\end{table}

\textsc{CoLA} reports accuracy rather than Matthews correlation so that all five tasks share a metric on the same scale, making the cross-task average meaningful.

\subsection{Methods compared}
At every rank $r \in \{4, 8, 16, 32, 64\}$ we train four parameter-efficient methods: \lora{} \citep{hu2022lora}, \textsc{DoRA} \citep{liu2024dora}, \textsc{AdaLoRA} \citep{zhang2023adalora} (target rank $r$, initial rank $2r$), and \ours{}. The four methods share the same target modules, optimizer, schedule, batch size, sequence length, and number of epochs (Table~\ref{tab:roberta-common-hp}); they differ only in the per-method configuration listed in the next subsection. We additionally train Full FT --- the entire encoder plus classification head --- as an upper-bound reference; it is not a competing PEFT method, and its per-task hyperparameters are listed separately in Table~\ref{tab:roberta-fullft}. Each (method, task, rank) cell is run with three seeds; the points in Figure~\ref{fig:roberta} are across-seed means with standard-deviation error bars.

\subsection{Common training hyperparameters}
The following are shared across all methods, tasks, and ranks unless explicitly overridden below.

\begin{table}[ht]
\centering
\renewcommand{\arraystretch}{1.2}
\caption{Common training hyperparameters shared across all methods, tasks, and ranks.}
\label{tab:roberta-common-hp}
\begin{tabular}{@{}ll@{}}
\toprule
\textbf{Hyperparameter} & \textbf{Value} \\
\midrule
\multicolumn{2}{@{}l}{\emph{Optimization}} \\
\quad Optimizer                 & AdamW ($\beta_1{=}0.9$, $\beta_2{=}0.999$, $\epsilon{=}10^{-8}$) \\
\quad Weight decay              & $0.01$ \\
\quad Max gradient norm         & $1.0$ \\
\addlinespace[2pt]
\multicolumn{2}{@{}l}{\emph{Learning-rate schedule}} \\
\quad Scheduler                 & cosine \\
\quad Warmup ratio              & $0.02$ \\
\addlinespace[2pt]
\multicolumn{2}{@{}l}{\emph{Batch and sequence}} \\
\quad Per-device batch size     & $8$ \\
\quad Gradient accumulation     & $2$ \\
\quad Effective batch size      & $16$ \\
\quad Max sequence length       & $512$ \\
\addlinespace[2pt]
\multicolumn{2}{@{}l}{\emph{Precision and hardware}} \\
\quad Precision                 & bfloat16 mixed precision \\
\quad GPUs per run              & $1$ \\
\addlinespace[2pt]
\multicolumn{2}{@{}l}{\emph{Adapter and training duration}} \\
\quad \lora{} dropout            & $0.0$ \\
\quad Epochs (adapter methods)   & $15$ \\
\quad Epochs (Full FT)           & $5$ on \textsc{MNLI}/\textsc{QNLI}/\textsc{SST-2}; $10$ on \textsc{CoLA}/\textsc{MRPC} \\
\bottomrule
\end{tabular}
\end{table}

\subsection{Method-specific hyperparameters}

\paragraph{Rank-scaled learning rate (\lora{}, \textsc{DoRA}, and the \lora{} branch of \ours{}).}
For these three methods we scale the learning rate with the adapter rank following \citet{shuttleworth2025lora}.

\paragraph{\textsc{AdaLoRA}.}
We set the initial rank to $\mathrm{init}_r = 2r$ and prune in a single step to the target rank $r$. The pruning step is chosen per task, roughly in proportion to the training-set size, giving $37\text{k}$ for \textsc{MNLI}, $10\text{k}$ for \textsc{QNLI}, $6\text{k}$ for \textsc{SST-2}, $800$ for \textsc{CoLA}, and $350$ for \textsc{MRPC} (so $t_{\mathrm{init}} = t_{\mathrm{final}}$). The remaining knobs follow the PEFT defaults: $\Delta_t = 10$, $\alpha = 2r$, and orthogonality-regularization weight $0.5$. The learning rate is held at $8\!\times\!10^{-4}$ for every task and rank.

\paragraph{\ours{} gate hyperparameters.}
The gate parameters $(\WW_g, \bb_g)$ are trained jointly with the \lora{} factors $(\AA, \BB)$. The \lora{} path uses the rank-scaled learning rate, and the gate uses a separate, larger one. We initialize $\WW_g$ with the PEFT default (scaled orthogonal initialization) and tune the gate bias $\bb_g$ and gate learning rate per task; ties on validation accuracy are broken in favor of the configuration with lower forgetting.

\begin{table}[ht]
\centering
\renewcommand{\arraystretch}{1.2}
\caption{Per-task \ours{} gate hyperparameters.}
\label{tab:roberta-gate-hp}
\begin{tabular}{@{}lll@{}}
\toprule
\textbf{Task} & \textbf{Gate bias init} $\bb_g$ & \textbf{Gate LR} \\
\midrule
\textsc{MNLI}  & $-7\cdot\mathbf{1}$ (uniform across ranks) & $1\!\times\!10^{-4}$ \\
\textsc{SST-2} & $-7\cdot\mathbf{1}$                          & $1\!\times\!10^{-4}$ \\
\textsc{QNLI}  & $-7\cdot\mathbf{1}$                          & $1\!\times\!10^{-4}$ \\
\textsc{CoLA}  & $-3$ at $r\in\{4,16,32\}$;\; $-7$ at $r\in\{8,64\}$ & $1\!\times\!10^{-3}$ \\
\textsc{MRPC}  & $-3$ at $r\in\{4,16\}$;\; $-7$ at $r\in\{8,32\}$;\; $-5$ at $r{=}64$ & $1\!\times\!10^{-3}$ \\
\bottomrule
\end{tabular}
\end{table}

\textsc{CoLA} and \textsc{MRPC} use a larger gate learning rate because their training sets are relatively small. With $\mathrm{gate\_lr}=10^{-4}$, the gates remain close to their near-closed initialization, leading to under-training of the \lora{} path and reduced accuracy. Increasing the gate learning rate to $10^{-3}$, together with a less negative bias, allows the gates to activate more readily and yields competitive performance. In contrast, for \textsc{MNLI}, \textsc{QNLI}, and \textsc{SST-2}, the smaller gate learning rate is sufficient and results in lower forgetting. A summary is provided in Table~\ref{tab:roberta-gate-hp}.

\paragraph{Full FT (reference upper bound).}
No adapter is used; every encoder parameter and the classification head are trained jointly. All other training hyperparameters match the adapter runs. Per-task LR and epoch count can be found in Table \ref{tab:roberta-fullft}.

\begin{table}[ht]
\centering
\renewcommand{\arraystretch}{1.2}
\caption{Full FT per-task hyperparameters.}
\label{tab:roberta-fullft}
\begin{tabular}{@{}lll@{}}
\toprule
\textbf{Task} & \textbf{Learning rate} & \textbf{Epochs} \\
\midrule
\textsc{MNLI}  & $1\!\times\!10^{-5}$ & 5  \\
\textsc{QNLI}  & $1\!\times\!10^{-5}$ & 5  \\
\textsc{SST-2} & $1\!\times\!10^{-5}$ & 5  \\
\textsc{CoLA}  & $2\!\times\!10^{-5}$ & 10 \\
\textsc{MRPC}  & $2\!\times\!10^{-5}$ & 10 \\
\bottomrule
\end{tabular}
\end{table}

\subsection{Evaluation protocol}

\paragraph{Best-checkpoint selection.}
We go over all the checkpoints we saved during training and pick the one with the best fine-tuning accuracy. The forgetting accuracy is also measured on this checkpoint. 

\paragraph{Fine-tuning score (y-axis of Figure~\ref{fig:roberta}).}
For each (method, rank, seed), the fine-tuning score is the unweighted mean of the five per-task validation accuracies,
\begin{equation}
    \text{FT score} = \tfrac{1}{5}\bigl(\text{\textsc{MNLI}}_{\text{acc}} + \text{\textsc{SST-2}}_{\text{acc}} + \text{\textsc{CoLA}}_{\text{acc}} + \text{\textsc{QNLI}}_{\text{acc}} + \text{\textsc{MRPC}}_{\text{acc}}\bigr).
\end{equation}

\paragraph{Forgetting score (x-axis of Figure~\ref{fig:roberta}).}
We measure forgetting as masked-LM perplexity on three held-out corpora. After fine-tuning, the encoder is paired with the original \roberta{} MLM head and evaluated on \textsc{WikiText-2} (HuggingFace \texttt{wikitext}, \texttt{wikitext-2-raw-v1}, validation split), \textsc{BookCorpusOpen} (\texttt{bookcorpusopen}), and \textsc{CC-News} (\texttt{cc\_news}); for the latter two we sample $10\text{k}$ sentences with a fixed seed. The figure plots the average of the three perplexities on a log scale. For reference, the pre-trained checkpoint scores $7.64$, $5.94$, and $5.63$ on the three corpora ($\mathrm{ppl}_{\text{base}} \approx 6.40$ overall).

\subsection{Infrastructure}
Each run uses a single NVIDIA A100 (40GB) GPU. We implement experiments in Python 3.13.5 with PyTorch (CUDA 13), using the HuggingFace \texttt{transformers}, \texttt{peft}, and \texttt{datasets} libraries, as well as the \texttt{lm-evaluation-harness}.

\section{\llama{} and \mistral{} Experiments: Details and Hyperparameters}\label{sec:appendix-llm}

In this section, we provide the full setup for the LLM experiments reported in Section~\ref{sec:exp-llm}.

\subsection{Models and training data}

We fine-tune two open-weight $7$B causal language models on two instruction-tuning corpora; the pairing of base model, corpus, sample count, and prompt template are summarized in Table~\ref{tab:llm-tasks}.

\begin{table}[ht]
\centering
\renewcommand{\arraystretch}{1.2}
\caption{Base models and instruction-tuning corpora.}
\label{tab:llm-tasks}
\begin{tabular}{@{}llrl@{}}
\toprule
\textbf{Setting} & \textbf{Base model} & \textbf{\#\,training samples} & \textbf{Prompt template} \\
\midrule
Math & \llama{}~$2$-$7$B \citep{touvron2023llama} (\texttt{Llama-2-7b-hf}) & ${\sim}\,395\text{K}$ & Alpaca \\
Code & \mistral{}-$7$B \citep{jiang2023mistral7b} (\texttt{Mistral-7B-v0.1}) & ${\sim}\,110\text{K}$ & Magicoder \\
\bottomrule
\end{tabular}
\end{table}

The math corpus is \textsc{MetaMathQA} \citep{yu2024metamath}; the code corpus is \textsc{Magicoder-Evol-Instruct-110K} \citep{wei2023magicoder}. Loss is computed only on response tokens; the instruction prefix is masked. We hold out a $5\%$ random validation split (seed $42$) and train on the remaining $95\%$. We do not perform validation-loss-based early stopping; every saved checkpoint is evaluated on the downstream benchmarks (Section~\ref{subsec:llm-eval}).

\paragraph{Prompt templates.}
The math runs use the standard Alpaca template:
\begin{verbatim}
### Instruction:
{instruction}

### Response:
{output}
\end{verbatim}

The code runs use the Magicoder template:
\begin{verbatim}
You are an exceptionally intelligent coding assistant that consistently delivers
accurate and reliable responses to user instructions.

@@ Instruction
{instruction}

@@ Response
{output}
\end{verbatim}

\subsection{Common training hyperparameters}

The following are shared across all methods, ranks, and seeds for both \llama{} and \mistral{} runs unless explicitly overridden in the method-specific section below; values are listed in Table~\ref{tab:llm-common-hp}.

\begin{table}[ht]
\centering
\renewcommand{\arraystretch}{1.2}
\caption{Training hyperparameters shared across all methods, ranks, and seeds.}
\label{tab:llm-common-hp}
\begin{tabular}{@{}ll@{}}
\toprule
\textbf{Hyperparameter} & \textbf{Value} \\
\midrule
\multicolumn{2}{@{}l}{\emph{Optimization}} \\
\quad Optimizer & AdamW ($\beta_1{=}0.9$, $\beta_2{=}0.999$, $\varepsilon{=}10^{-8}$) \\
\quad Weight decay & $0.01$ (excluded from biases, LayerNorm, and gate parameters) \\
\quad Max gradient norm & $1.0$ \\
\addlinespace[2pt]
\multicolumn{2}{@{}l}{\emph{Learning-rate schedule}} \\
\quad Scheduler & cosine \\
\quad Warmup ratio & $0.02$ \\
\addlinespace[2pt]
\multicolumn{2}{@{}l}{\emph{Batch and sequence}} \\
\quad Effective batch size & $32$ sequences per optimizer step \\
\quad Max sequence length & $512$ (math) / $2048$ (code) \\
\addlinespace[2pt]
\multicolumn{2}{@{}l}{\emph{Precision and hardware}} \\
\quad Precision & bfloat16 \\
\quad Gradient checkpointing & enabled \\
\quad GPUs per run & $4 \times$ NVIDIA A100 (40\,GB), data-parallel \\
\addlinespace[2pt]
\multicolumn{2}{@{}l}{\emph{Training duration}} \\
\quad Epochs & $3$ \\
\quad Checkpoints saved & $2$ per epoch (cap $16$, oldest pruned) \\
\bottomrule
\end{tabular}
\end{table}

\subsection{Adapter configuration}

The four adapter methods (\lora{}, \textsc{DoRA}, \textsc{AdaLoRA}, and \ours{}) share the following baseline configuration.

\begin{itemize}
    \item \textbf{Target modules.} Every linear projection inside the decoder blocks: \texttt{q\_proj}, \texttt{k\_proj}, \texttt{v\_proj}, \texttt{o\_proj}, \texttt{gate\_proj}, \texttt{up\_proj}, and \texttt{down\_proj}. The LM head and embeddings are frozen.
    \item \textbf{Bias.} Not trained.
    \item \textbf{Adapter dropout.} $0.0$.
    \item \textbf{Scaling.} $\alpha = 2r$.
    \item \textbf{Initialization.} The \lora{} factor $\AA$ is Kaiming-uniform, $\BB = \zeroVec$. 
\end{itemize}

\subsection{Method-specific hyperparameters}

\paragraph{\lora{}.}
Standard \lora{} \citep{hu2022lora} as implemented in HuggingFace \texttt{peft}: for each pre-trained weight $\WW \in \R^{d_y \times d_x}$ in the target set, we learn $\bDelta = \frac{\alpha}{r}\,\AA\BB$ with $\AA \in \R^{d_y \times r}$ and $\BB \in \R^{r \times d_x}$, matching the notation of Section~\ref{sec:method}.

\paragraph{\textsc{AdaLoRA}.}
We use the original recipe of \citet{zhang2023adalora} via HuggingFace \texttt{peft}, with initial rank $\mathrm{init}_r = 1.5\,r$ and the budget-reduction schedule of the paper's \textsc{SQuAD} setting (warm-up to $t_{\mathrm{init}} \approx 17\%$ of total steps, pruning ending at $t_{\mathrm{final}} \approx 33\%$, $\Delta_t = 10$, $\beta_1 = \beta_2 = 0.85$); per-setting step counts are given in Table~\ref{tab:adalora-schedule}. We set the orthogonality-regularization weight to $0.1$, again following the original paper, since the PEFT default of $0.5$ dominated the training loss in pilot runs. All other adapter settings ($\alpha = 2r$, dropout, target modules) match \lora{}. The learning rate is $2 \times 10^{-4}$ for $r \in \{16, 32, 64, 128\}$ and $1 \times 10^{-4}$ for $r = 256$ to avoid divergence at $\mathrm{init}_r = 384$.

\begin{table}[ht]
\centering
\renewcommand{\arraystretch}{1.2}
\caption{\textsc{AdaLoRA} budget-reduction schedule per setting. All ranks within a setting share the same total step count.}
\label{tab:adalora-schedule}
\begin{tabular}{@{}lrrr@{}}
\toprule
\textbf{Setting} & \textbf{Total steps} & $\bm{t_{\mathrm{init}}}$ & $\bm{t_{\mathrm{final}}}$ \\
\midrule
\llama{}~$2$~$7$B / \textsc{MetaMathQA} & $8\,796$ & $1\,574$ & $3\,055$ \\
\mistral{}~$7$B / \textsc{Magicoder} & $2\,478$ & $420$ & $820$ \\
\bottomrule
\end{tabular}
\end{table}

\paragraph{Full FT.}
All transformer parameters are unfrozen ($\approx 6.74$B for \llama{}~$2$~$7$B and $\approx 7.24$B for \mistral{}~$7$B), including the LM head and the input embeddings. Two learning rates were swept ($10^{-5}$ and $5 \times 10^{-5}$); we report $10^{-5}$ in the main figure ($5 \times 10^{-5}$ was unstable on math). All other settings (optimizer, schedule, batch size, epochs, precision, gradient checkpointing) match the adapter runs.

\paragraph{\ours{} gate hyperparameters.}
For every adapted layer, an input-dependent rank gate is attached to the \lora{} branch:
\begin{equation}
    \bDelta(\xx)\,\xx \;=\; \frac{\alpha}{r}\, \AA\,\GG(\xx)\,\BB\,\xx,
    \qquad
    g(\xx) \;=\; \sigma\!\rb{\WW_g\,\xx + \bb_g} \in (0,1)^r,
\end{equation}
with $\WW_g \in \R^{r \times d_x}$, matching the notation of Section~\ref{sec:method}. The gate adds $r\,d_x + r$ parameters per adapted layer. The remaining gate-specific hyperparameters (bias and weight initialization, gate normalization, gate learning rate, target modules) are listed in Table~\ref{tab:llm-gate-hp}.

\begin{table}[ht]
\centering
\renewcommand{\arraystretch}{1.2}
\caption{\ours{} gate hyperparameters used for both \llama{} and \mistral{}.}
\label{tab:llm-gate-hp}
\begin{tabular}{@{}ll@{}}
\toprule
\textbf{Gate hyperparameter} & \textbf{Value} \\
\midrule
Bias init $\bb_g$ & $-3.0$ ($\sigma(-3) \approx 0.047$, gates start nearly closed) \\
Weight init $\WW_g$ & Kaiming-uniform \\
Gate normalization & off \\
Gate learning rate & $10^{-3}$ (separate AdamW group; weight decay disabled) \\
Targets & identical to \lora{} \\
\bottomrule
\end{tabular}
\end{table}

\subsection{Per-rank learning rates}

For every method we report the learning rate that gave the highest target-task accuracy in our sweep over $\{10^{-4}, 2\!\times\!10^{-4}, 4\!\times\!10^{-4}\}$ for adapter methods, and $\{10^{-5}, 5\!\times\!10^{-5}, 10^{-4}\}$ for Full FT. The effective batch size is held at $32$ for every run; per-device batch and gradient-accumulation count are adjusted at the largest ranks (down to per-device $2 \times 16$ for \mistral{} \ours{} at $r{=}256$) to fit GPU memory. The selected learning rates for every (method, rank, setting) cell are summarized in Table~\ref{tab:llm-runs}.

\begin{table}[ht]
\centering
\renewcommand{\arraystretch}{1.2}
\caption{Per-run learning rates for both fine-tuning settings.}
\label{tab:llm-runs}
\begin{tabular}{@{}llcc@{}}
\toprule
\textbf{Method} & \textbf{Rank $r$} & \textbf{\llama{}~$2$~$7$B (math)} & \textbf{\mistral{}~$7$B (code)} \\
\midrule
\lora{} & $16$  & $2 \times 10^{-4}$ & $2 \times 10^{-4}$ \\
\lora{} & $32$  & $2 \times 10^{-4}$ & $2 \times 10^{-4}$ \\
\lora{} & $64$  & $2 \times 10^{-4}$ & $1 \times 10^{-4}$ \\
\lora{} & $128$ & $2 \times 10^{-4}$ & $2 \times 10^{-4}$ \\
\lora{} & $256$ & $1 \times 10^{-4}$ & $1 \times 10^{-4}$ \\
\addlinespace[2pt]
\textsc{DoRA} & $16$  & $2 \times 10^{-4}$ & $2 \times 10^{-4}$ \\
\textsc{DoRA} & $32$  & $2 \times 10^{-4}$ & $2 \times 10^{-4}$ \\
\textsc{DoRA} & $64$  & $2 \times 10^{-4}$ & $2 \times 10^{-4}$ \\
\textsc{DoRA} & $128$ & $2 \times 10^{-4}$ & $2 \times 10^{-4}$ \\
\textsc{DoRA} & $256$ & $1 \times 10^{-4}$ & $1 \times 10^{-4}$ \\
\addlinespace[2pt]
\textsc{AdaLoRA} & $16$  & $2 \times 10^{-4}$ & $2 \times 10^{-4}$ \\
\textsc{AdaLoRA} & $32$  & $2 \times 10^{-4}$ & $2 \times 10^{-4}$ \\
\textsc{AdaLoRA} & $64$  & $2 \times 10^{-4}$ & $2 \times 10^{-4}$ \\
\textsc{AdaLoRA} & $128$ & $2 \times 10^{-4}$ & $2 \times 10^{-4}$ \\
\textsc{AdaLoRA} & $256$ & $1 \times 10^{-4}$ & $1 \times 10^{-4}$ \\
\addlinespace[2pt]
\ours{} & $16$  & $2 \times 10^{-4}$ & $2 \times 10^{-4}$ \\
\ours{} & $32$  & $2 \times 10^{-4}$ & $2 \times 10^{-4}$ \\
\ours{} & $64$  & $2 \times 10^{-4}$ & $2 \times 10^{-4}$ \\
\ours{} & $128$ & $2 \times 10^{-4}$ & $2 \times 10^{-4}$ \\
\ours{} & $256$ & $1 \times 10^{-4}$ & $1 \times 10^{-4}$ \\
\addlinespace[2pt]
Full FT & no adapter & $1 \times 10^{-5}$ & $1 \times 10^{-5}$ \\
\bottomrule
\end{tabular}
\end{table}

\subsection{Evaluation protocol}\label{subsec:llm-eval}

All evaluation runs use a single A100 GPU with bfloat16 inference. The adapter is \emph{not} merged into the base weights: the PEFT model is loaded in evaluation mode with all adapter parameters and (for \ours{}) the gate parameters restored from disk.

\paragraph{Target-task evaluation.}
\begin{itemize}
    \item \textbf{Math (\llama{}~$2$~$7$B / \textsc{MetaMathQA})}: \textsc{GSM8K}, $5$-shot, \texttt{exact\_match,strict-match}.
    \item \textbf{Code (\mistral{}~$7$B / \textsc{Magicoder})}: \textsc{HumanEval} pass$@1$ with greedy decoding, one sample per problem ($164$ problems).
\end{itemize}

\paragraph{Forgetting suite (both settings).}
The retention score reported in the main figure is the unweighted mean accuracy across the $14$ benchmarks listed in Table~\ref{tab:llm-forgetting-suite}, all evaluated $5$-shot under the same \texttt{lm-eval} configuration with \texttt{max\_length=512}. The same suite is used for both math and code runs.

\begin{table}[ht]
\centering
\renewcommand{\arraystretch}{1.2}
\caption{Forgetting (retention) suite: $14$ benchmarks evaluated $5$-shot under the same \texttt{lm-eval} configuration for both \llama{} (math) and \mistral{} (code) runs.}
\label{tab:llm-forgetting-suite}
\begin{tabular}{@{}lll@{}}
\toprule
\textbf{Benchmark} & \textbf{Category} & \textbf{Metric} \\
\midrule
\textsc{HellaSwag}~\citep{hellaswag}                & Commonsense reasoning  & \texttt{acc\_norm} \\
\textsc{WinoGrande}~\citep{winogrande}              & Commonsense reasoning  & \texttt{acc} \\
\textsc{ARC-Challenge}~\citep{allenai:arc}            & Science QA             & \texttt{acc\_norm} \\
\textsc{ARC-Easy}~\citep{allenai:arc}                 & Science QA             & \texttt{acc\_norm} \\
\textsc{LAMBADA-OpenAI}~\citep{lambda_openai}           & Language modeling      & \texttt{acc} (and perplexity) \\
\textsc{BoolQ}~\citep{boolq}                    & Reading comprehension  & \texttt{acc} \\
\textsc{PIQA}~\citep{piqa}                     & Physical commonsense   & \texttt{acc\_norm} \\
\textsc{TriviaQA}~\citep{2017arXivtriviaqa}                 & Open-domain QA         & \texttt{acc} \\
\textsc{OpenBookQA}~\citep{OpenBookQA}               & Science QA             & \texttt{acc\_norm} \\
\textsc{SciQ}~\citep{SciQ}                     & Science QA             & \texttt{acc} \\
\textsc{MMLU} ($57$-task aggregate)~\citep{mmlu1} & Multitask knowledge   & \texttt{acc} \\
\textsc{MedQA-4options}~\citep{jin2020disease}           & Medical QA             & \texttt{acc} \\
\textsc{CommonsenseQA}~\citep{talmor-etal-2019-commonsenseqa}            & Commonsense reasoning  & \texttt{acc} \\
\textsc{NaturalQuestions-Open}    & Open-domain QA         & \texttt{acc} \\
\bottomrule
\end{tabular}
\end{table}

\clearpage
\section{Checkpoint Dynamics: Retention and Fine-Tuning Accuracy}\label{sec:appendix-retention-over-time}

\subsection{\llama{}~2~7B / \textsc{MetaMathQA}}\label{sec:appendix-checkpoint-llama}

\begin{figure}[t]
\centering
\begin{subfigure}[t]{0.48\textwidth}
\centering
\includegraphics[width=\textwidth]{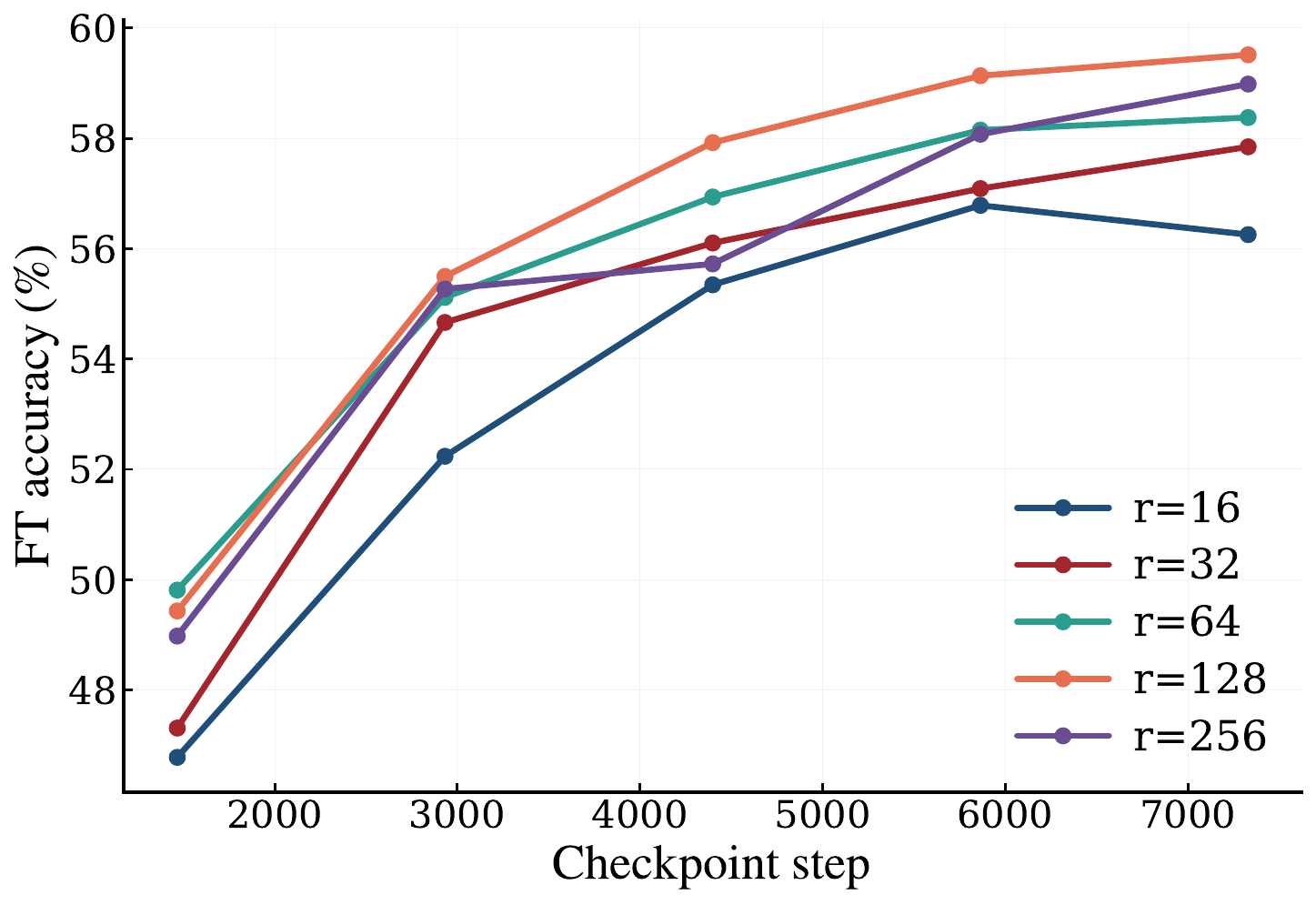}
\caption{\textbf{\ours{}:} \textsc{GSM8K} accuracy improves over training across ranks.}
\label{fig:llama-ft-dial}
\end{subfigure}
\hfill
\begin{subfigure}[t]{0.48\textwidth}
\centering
\includegraphics[width=\textwidth]{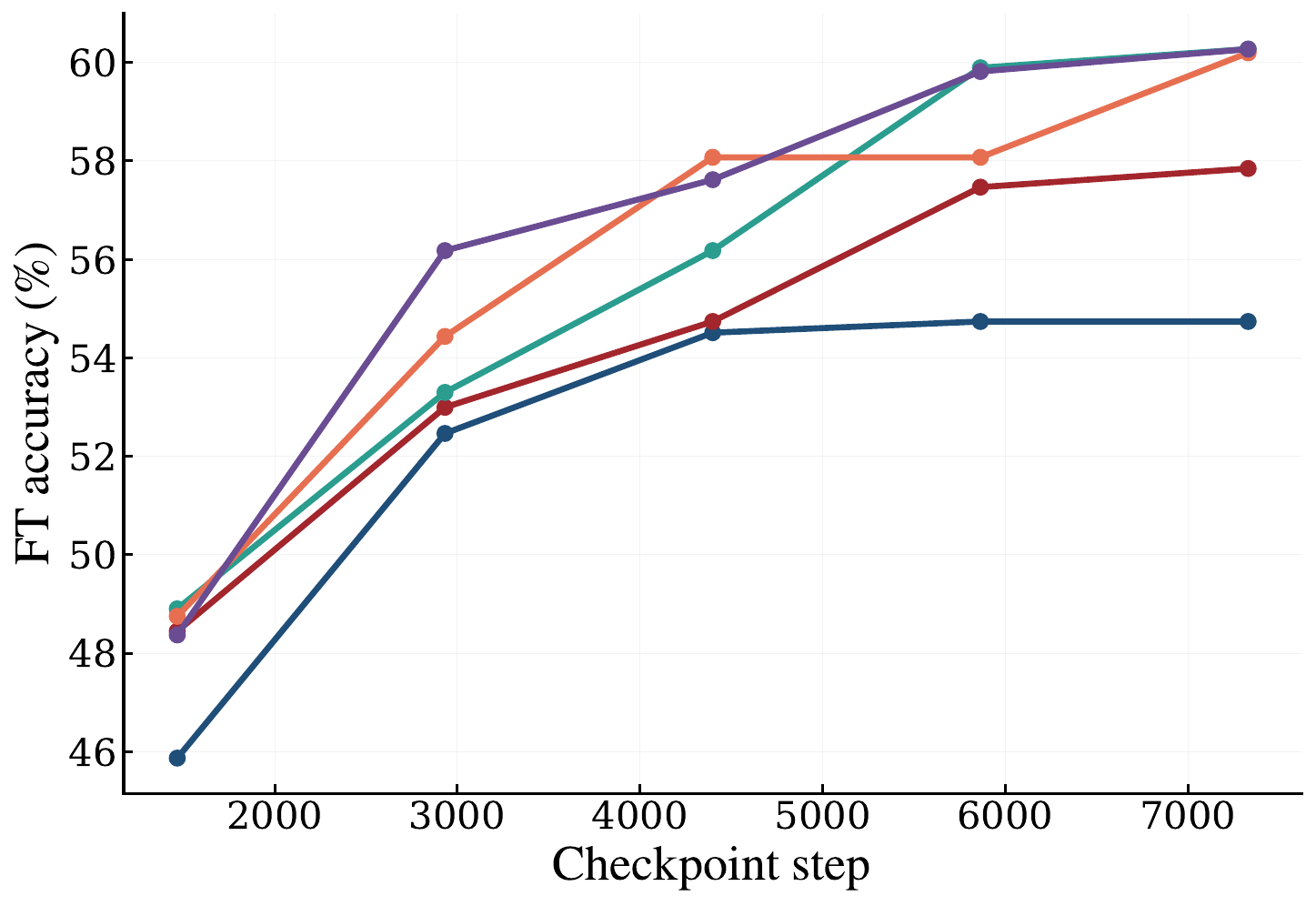}
\caption{\textbf{\lora{}:} \textsc{GSM8K} accuracy also improves, but retention degrades at larger ranks.}
\label{fig:llama-ft-lora}
\end{subfigure}
\caption{\textbf{Fine-tuning accuracy over training checkpoints for \lora{} and \ours{} on \llama{}~$2$~$7$B / \textsc{MetaMathQA}.} Each curve reports \textsc{GSM8K} accuracy at every saved checkpoint, for adapter ranks $r \in \{16, 32, 64, 128, 256\}$.}
\label{fig:llama-ft-over-time}
\end{figure}

Section~\ref{sec:retention_over_time} reports the checkpoint-level retention curves for the \llama{}~$2$~$7$B / \textsc{MetaMathQA} run. Figure~\ref{fig:llama-ft-over-time} shows the corresponding fine-tuning accuracy curves. Both \ours{} and \lora{} improve on \textsc{GSM8K} over training, but the retention curves in Figure~\ref{fig:retention-over-time} show that \ours{} achieves this without the rank-dependent retention drift observed for \lora{}.

\subsection{\mistral{}~7B / \textsc{Magicoder}}\label{sec:appendix-retention-mistral}

\begin{figure}[t]
\centering
\begin{subfigure}[t]{0.48\textwidth}
\centering
\includegraphics[width=\textwidth]{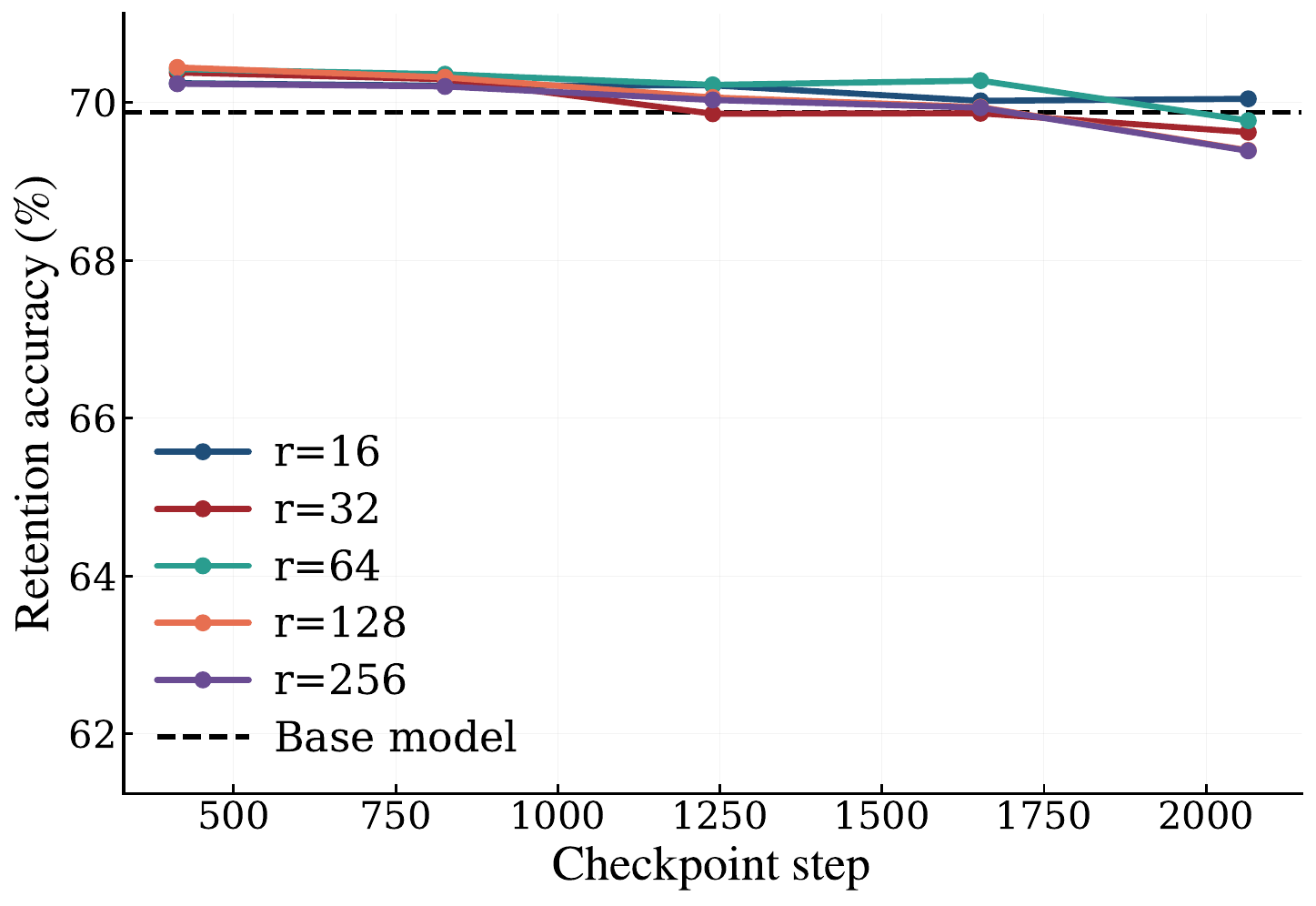}
\caption{\textbf{\ours{}:} retention stays close to the base model across ranks and training checkpoints.}
\label{fig:mistral-retention-dial}
\end{subfigure}
\hfill
\begin{subfigure}[t]{0.48\textwidth}
\centering
\includegraphics[width=\textwidth]{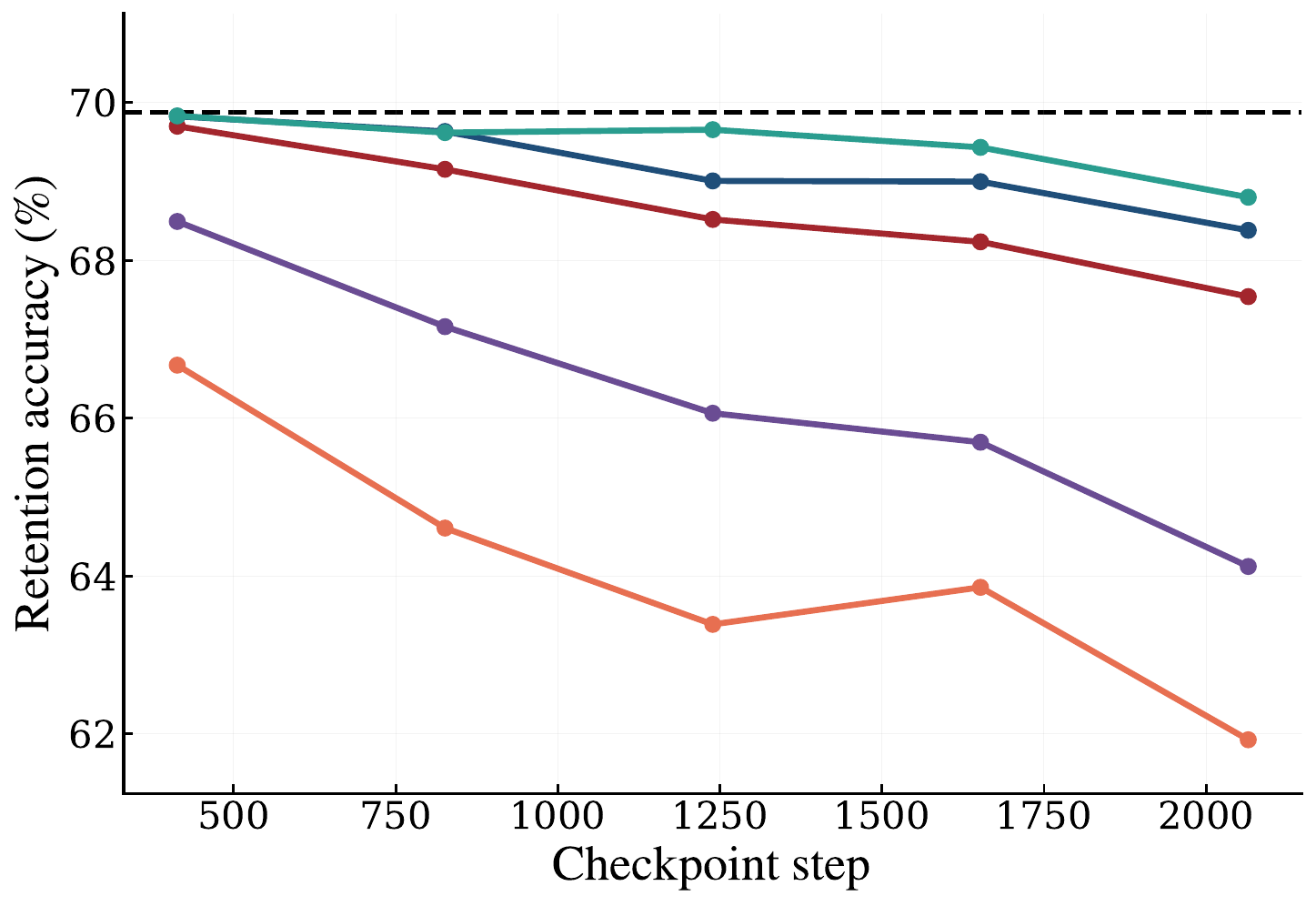}
\caption{\textbf{\lora{}:} retention degrades over training, with larger drops at higher ranks.}
\label{fig:mistral-retention-lora}
\end{subfigure}
\caption{\textbf{Retention over training checkpoints for \lora{} and \ours{} on \mistral{}~$7$B / \textsc{Magicoder}.} Each curve tracks the unweighted mean accuracy across the $14$-benchmark retention suite at every saved checkpoint, for adapter ranks $r \in \{16, 32, 64, 128, 256\}$.}
\label{fig:mistral-retention-over-time}
\end{figure}

\begin{figure}[t]
\centering
\begin{subfigure}[t]{0.48\textwidth}
\centering
\includegraphics[width=\textwidth]{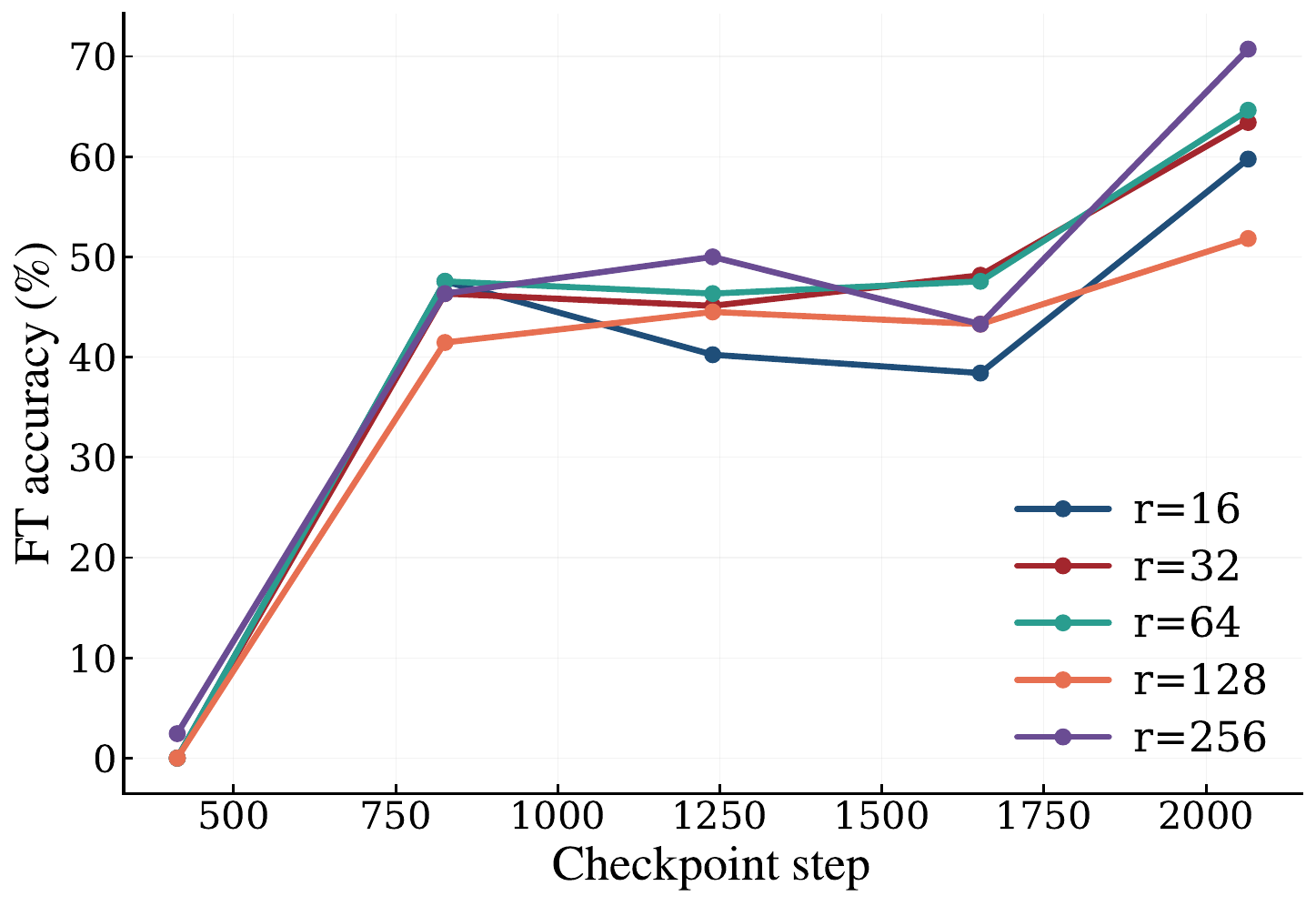}
\caption{\textbf{\ours{}:} target-task accuracy improves over training and remains competitive across ranks.}
\label{fig:mistral-ft-dial}
\end{subfigure}
\hfill
\begin{subfigure}[t]{0.48\textwidth}
\centering
\includegraphics[width=\textwidth]{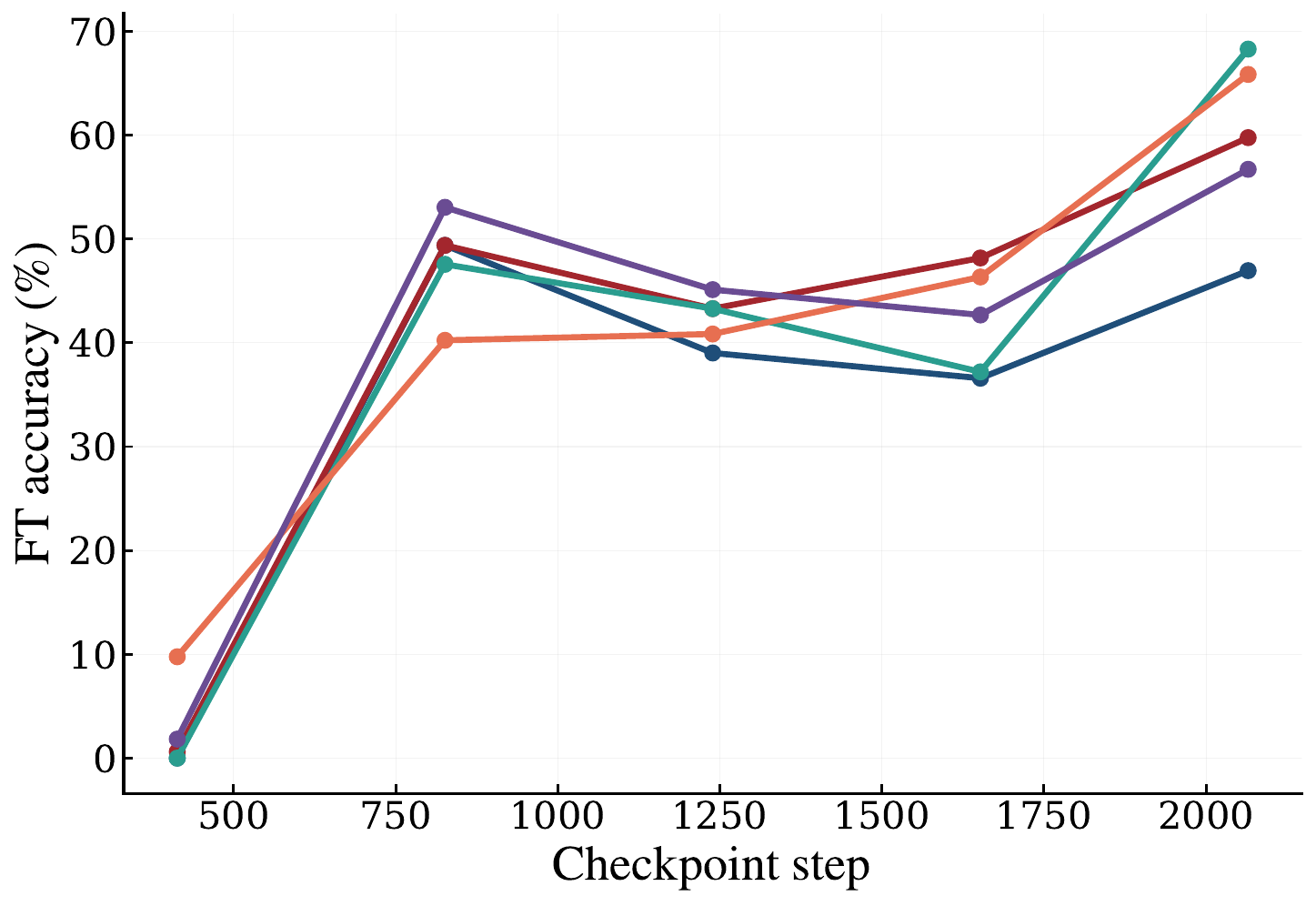}
\caption{\textbf{\lora{}:} target-task accuracy also improves, but with larger retention loss at high ranks.}
\label{fig:mistral-ft-lora}
\end{subfigure}
\caption{\textbf{Fine-tuning accuracy over training checkpoints for \lora{} and \ours{} on \mistral{}~$7$B / \textsc{Magicoder}.} Each curve reports \textsc{HumanEval} pass@$1$ at every saved checkpoint, for adapter ranks $r \in \{16, 32, 64, 128, 256\}$.}
\label{fig:mistral-ft-over-time}
\end{figure}

Figures~\ref{fig:mistral-retention-over-time} and~\ref{fig:mistral-ft-over-time} repeat the checkpoint-level analysis for the \mistral{}~$7$B / \textsc{Magicoder} code-generation run. The retention pattern mirrors the \llama{} math setting in Section~\ref{sec:retention_over_time}: \ours{} keeps retention close to the base model throughout training and across ranks, whereas \lora{} exhibits rank-dependent drift. At the same time, the fine-tuning accuracy curves show that this retention advantage does not come from under-training; \ours{} reaches comparable or stronger target-task accuracy at the final checkpoints.

\section{Interpretability: Additional Results}\label{sec:appendix-interpretability}

This appendix complements the gate-activation analysis of Section~\ref{sec:interpretability}, which focused on \llama{}~$2$~$7$B at rank $r{=}32$. We report (i) additional \llama{}~$2$~$7$B results across ranks and modules, and (ii) the corresponding analysis on \mistral{}~$7$B fine-tuned for code generation. The setup, evaluation domains, and statistics shown (sigmoid gate values aggregated over held-out inputs) are identical to Section~\ref{sec:interpretability}; only the model, rank, or module under inspection differs.

\subsection{\llama{}~2~7B: additional ranks and modules}\label{sec:appendix-llama-extra}

We extend the rank-$32$ analysis of Section~\ref{sec:interpretability} along two axes. First, we recompute the domain-conditional gate distributions (math, code, general text) at additional adapter ranks to check that the depth-band pattern reported in the main text is not specific to $r{=}32$. Second, we extend the per-module breakdown to the remaining linear projections in the decoder (\texttt{q\_proj}, \texttt{k\_proj}, \texttt{o\_proj}, \texttt{gate\_proj}), so the $v_{\mathrm{proj}}$ / $\mathrm{up}_{\mathrm{proj}}$ / $\mathrm{down}_{\mathrm{proj}}$ comparison in Figure~\ref{fig:llama-gate-modules} can be read in context of the full attention and MLP stacks. The qualitative picture is the same as in the main text: math inputs reach the highest gate values, the activation is concentrated in mid-to-late layers, and within those layers, the MLP path (especially $\mathrm{up}_{\mathrm{proj}}$) carries most of the active capacity.

\begin{figure}[t]
\centering
\includegraphics[width=\textwidth]{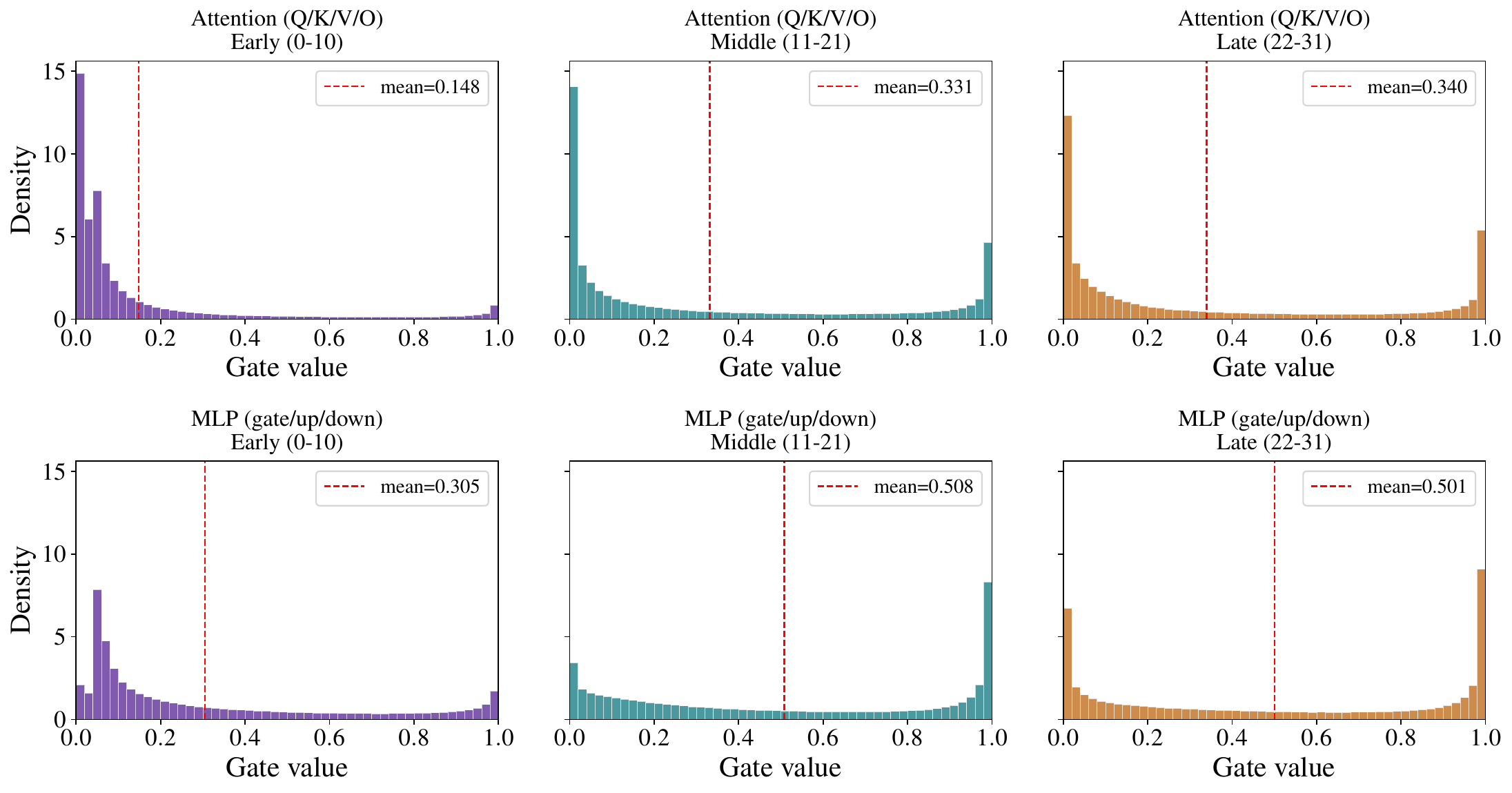}
\caption{\textbf{Gate distributions by module family and depth on \llama{} math inputs.} We aggregate gate values from the rank-$32$ \llama{} math adapter over held-out \textsc{GSM8K} examples. Attention denotes \texttt{q\_proj}, \texttt{k\_proj}, \texttt{v\_proj}, and \texttt{o\_proj}; MLP denotes \texttt{gate\_proj}, \texttt{up\_proj}, and \texttt{down\_proj}. Gates open most strongly in the middle and late layers, especially in the MLP stack.}
\label{fig:appendix-llama-module-layer-math}
\end{figure}

\begin{figure}[t]
\centering
\includegraphics[width=\textwidth]{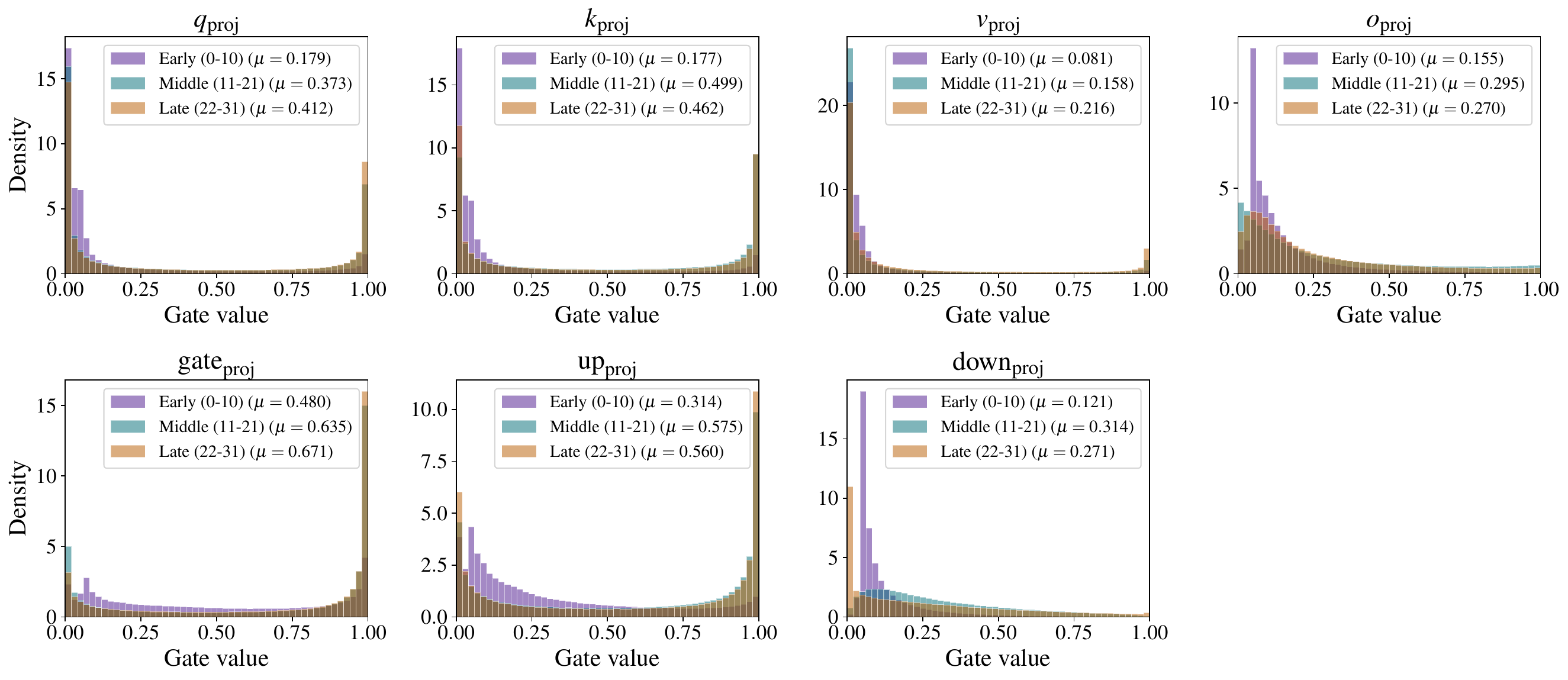}
\caption{\textbf{Gate distributions by projection on \llama{} math inputs.} We show gate-value histograms for all adapted projections of the rank-$32$ \llama{} math adapter, grouped by layer-depth band. The MLP \texttt{gate\_proj} and \texttt{up\_proj} are the most active, while \texttt{v\_proj} and early \texttt{down\_proj} remain close to closed.}
\label{fig:appendix-llama-per-module-math}
\end{figure}

Figure~\ref{fig:appendix-llama-module-layer-math} first aggregates the adapted projections into attention (\texttt{q\_proj}, \texttt{k\_proj}, \texttt{v\_proj}, \texttt{o\_proj}) and MLP (\texttt{gate\_proj}, \texttt{up\_proj}, \texttt{down\_proj}) groups on held-out \textsc{GSM8K} inputs. Both groups become more active in middle and late layers, but the MLP gates are consistently more open: the mean gate value rises from about $0.31$ in early layers to about $0.50$ in middle and late layers, compared with about $0.15$ to $0.34$ for attention. Figure~\ref{fig:appendix-llama-per-module-math} breaks this down by projection. The largest mean activations occur in \texttt{gate\_proj}, \texttt{up\_proj}, and the attention key/query projections in later layers, while \texttt{v\_proj} and early \texttt{down\_proj} remain mostly closed. This finer view supports the main-text conclusion that \ours{} does not simply open all ranks uniformly: it concentrates capacity in specific depths and projections.

\subsection{\mistral{}~7B fine-tuned on code}\label{sec:appendix-mistral-extra}

\begin{figure}[t]
\centering
\includegraphics[width=\textwidth]{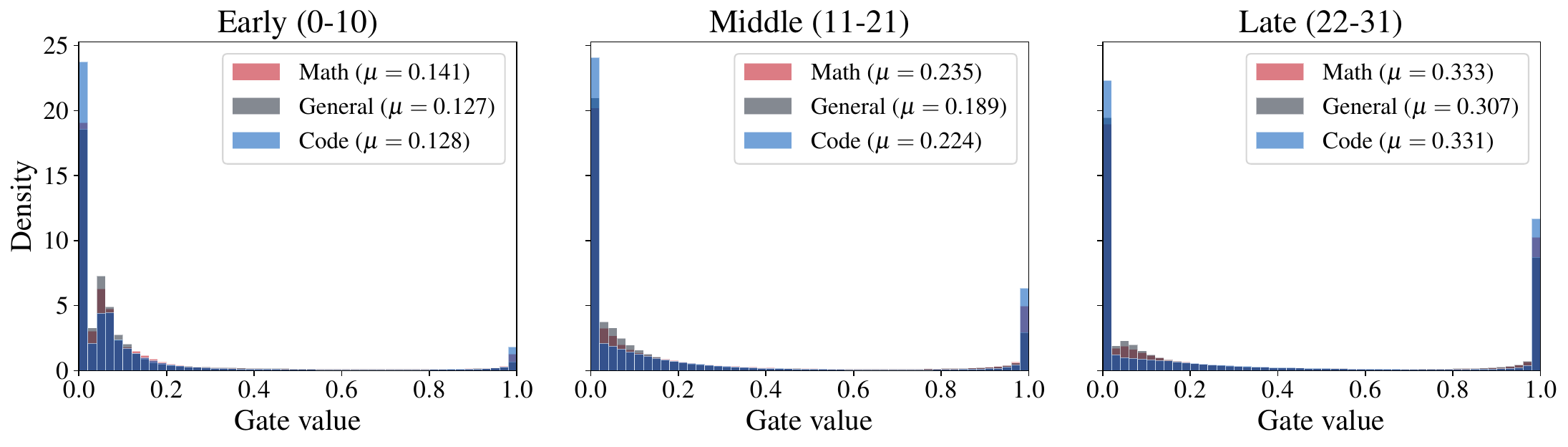}
\caption{\textbf{Gate activation histograms across domains for the \mistral{} code adapter.} We analyze the rank-$32$ \mistral{} adapter fine-tuned on \textsc{Magicoder}. Unlike the \llama{} math adapter, where math is clearly separated from code, the \mistral{} code adapter opens gates to a similar degree on held-out code and math inputs, while general text remains lower. Code nevertheless has the largest concentration of fully open gates, even when math has a comparable or slightly higher mean gate value.}
\label{fig:appendix-mistral-domain-code}
\end{figure}

We repeat the diagnostics of Section~\ref{sec:interpretability} on \mistral{}~$7$B fine-tuned on \textsc{Magicoder}, using held-out code inputs as the fine-tuning domain and math/general text as contrast domains. Figure~\ref{fig:appendix-mistral-domain-code} shows the main domain-level difference from the \llama{} math adapter. On \llama{}, math inputs are clearly more activated than code, with general text lowest. On \mistral{} fine-tuned for code, code and math are nearly indistinguishable in mean gate value: early layers give $0.128$ for code and $0.141$ for math, middle layers give $0.224$ and $0.235$, and late layers give $0.331$ and $0.333$. However, the histogram shape still reflects the fine-tuning domain: code inputs have the largest mass of gates that are fully open, even though math sometimes has a slightly larger mean because it has more partially open gates. General text remains lower, but the gap is smaller than in the \llama{} math setting. This suggests that the code adapter treats math and code as closely related structured-symbolic inputs, while still opening its highest-confidence gates most often on code.

\begin{figure}[t]
\centering
\includegraphics[width=\textwidth]{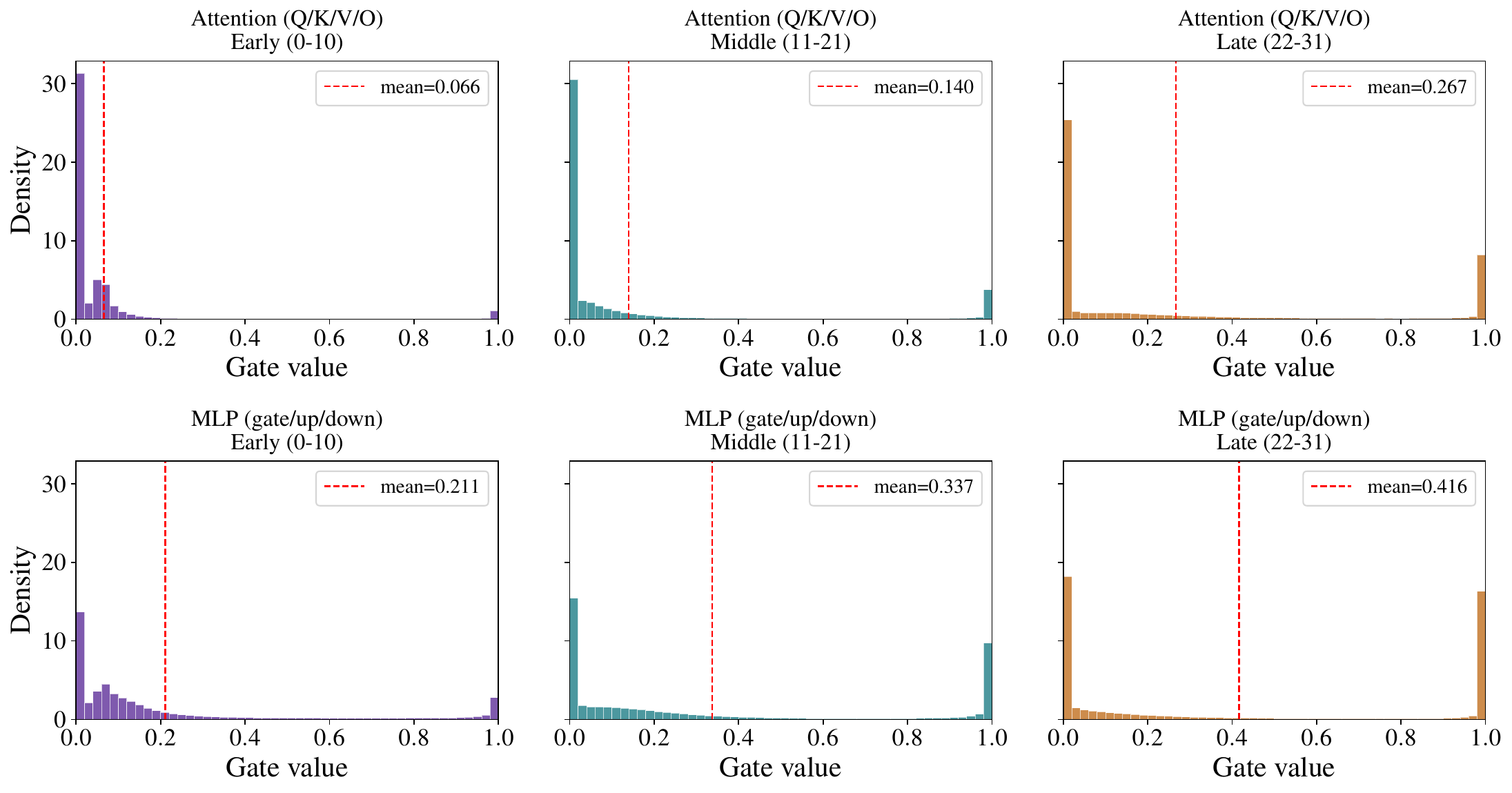}
\caption{\textbf{Gate distributions by module family and depth on \mistral{} code inputs.} Attention denotes \texttt{q\_proj}, \texttt{k\_proj}, \texttt{v\_proj}, and \texttt{o\_proj}; MLP denotes \texttt{gate\_proj}, \texttt{up\_proj}, and \texttt{down\_proj}. As in the \llama{} analysis, gates open most strongly in the MLP stack and in middle-to-late layers.}
\label{fig:appendix-mistral-module-layer-code}
\end{figure}

\begin{figure}[t]
\centering
\includegraphics[width=\textwidth]{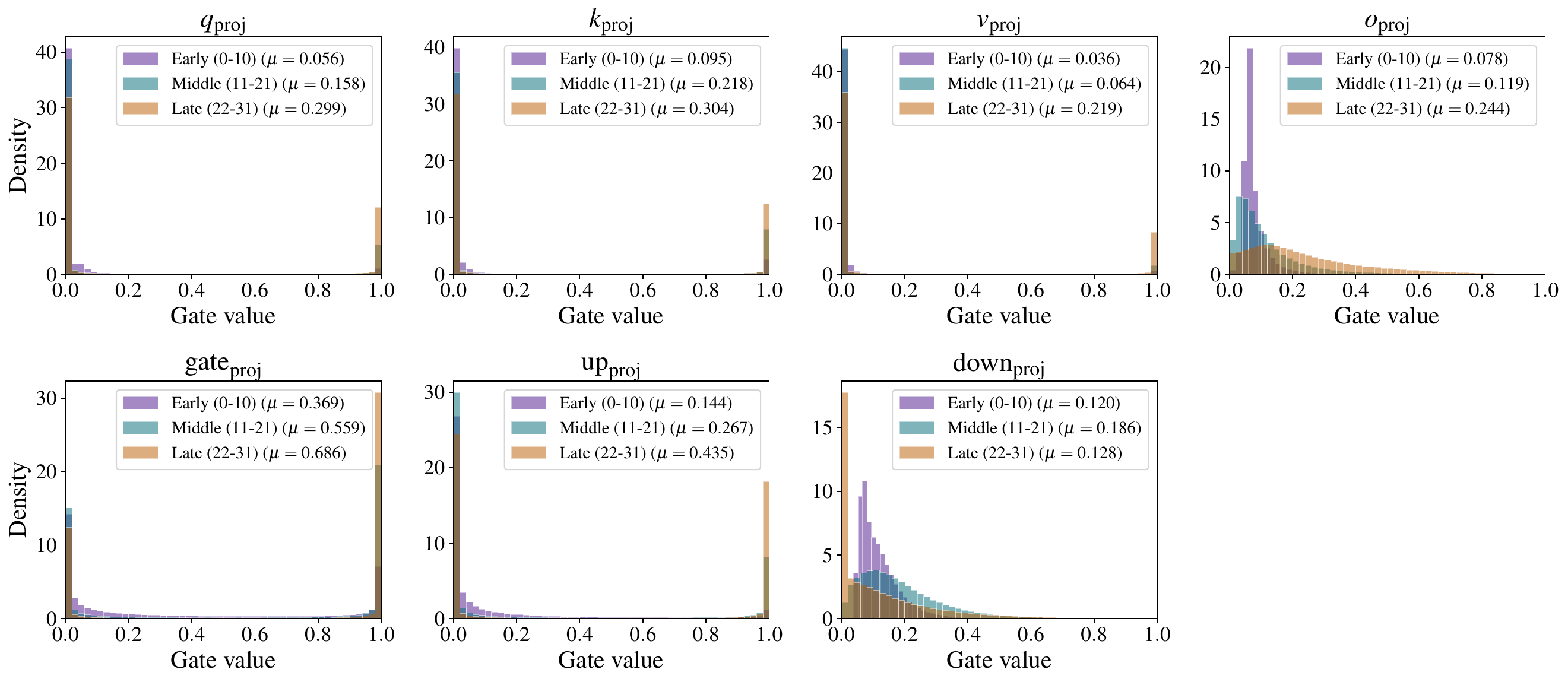}
\caption{\textbf{Gate distributions by projection on \mistral{} code inputs.} We show gate-value histograms for all adapted projections of the rank-$32$ \mistral{} code adapter, grouped by layer-depth band. The largest activations occur in the MLP \texttt{gate\_proj} and \texttt{up\_proj}, while several attention projections remain mostly closed until late layers.}
\label{fig:appendix-mistral-per-module-code}
\end{figure}

The module-level views in Figures~\ref{fig:appendix-mistral-module-layer-code} and~\ref{fig:appendix-mistral-per-module-code} show that the coarse organization is nevertheless similar to \llama{}. Gates open more in the middle and late layers than in the early layers, and the MLP stack is more active than the attention stack. Aggregating by module family, MLP mean activation rises from $0.211$ in early layers to $0.416$ in late layers, whereas attention rises from $0.066$ to $0.267$. The per-projection breakdown shows the strongest activations in \texttt{gate\_proj} and \texttt{up\_proj}, with \texttt{gate\_proj} reaching the largest late-layer mean ($\mu \approx 0.686$). Thus, while the domain selectivity differs from \llama{}---code and math open the gates almost equally on \mistral{}---the layer and module selectivity remains consistent.

\end{document}